\documentclass{article}

\usepackage[nonatbib, final]{neurips_2025}

\usepackage[utf8]{inputenc} % allow utf-8 input
\usepackage[T1]{fontenc}    % use 8-bit T1 fonts
\usepackage{hyperref}       % hyperlinks
\usepackage{url}            % simple URL typesetting
\usepackage{booktabs}       % professional-quality tables
\usepackage{amsfonts}       % blackboard math symbols
\usepackage{nicefrac}       % compact symbols for 1/2, etc.
\usepackage{microtype}      % microtypography
\usepackage{xcolor}         % colors

\usepackage{graphicx}
\usepackage{sidecap}
\usepackage{multicol}
\usepackage{amsmath}
\usepackage{orcidlink}
\usepackage{tikz}           % figures and pictures
\usepackage{appendix}       % supplementary information
\usepackage[sortcites, backend=biber, sorting=none, maxnames=99]{biblatex}
\usepackage{multirow}

\sidecaptionvpos{figure}{t}

\title{Explicitly Modeling Subcortical Vision with a Neuro-Inspired Front-End Improves CNN Robustness}

\author{%
  Lucas Piper\orcidlink{0009-0000-5963-1163}$^{1,2}$, Arlindo L. Oliveira\orcidlink{0000-0001-8638-5594}$^{1,2}$, Tiago Marques\orcidlink{0000-0002-8973-0549}$^{3,4}$ \\
  \\
  $^1$INESC-ID, Lisboa, Portugal \\
  $^2$Instituto Superior Técnico, Universidade de Lisboa, Lisboa, Portugal \\
  $^3$Breast Cancer Research Program, Champalimaud Foundation, Lisboa, Portugal \\
  $^4$Faculdade de Medicina de Lisboa, Universidade de Lisboa, Portugal \\
  \texttt{lucaspiper99@tecnico.ulisboa.pt} \\
}

\begin{document}

\maketitle

\begin{abstract}
    Convolutional neural networks (CNNs) trained on object recognition achieve high task performance but continue to exhibit vulnerability under a range of visual perturbations and out-of-domain images, when compared with biological vision. Prior work has demonstrated that coupling a standard CNN with a front-end (VOneBlock) that mimics the primate primary visual cortex (V1) can improve overall model robustness. Expanding on this, we introduce Early Vision Networks (EVNets), a new class of hybrid CNNs that combine the VOneBlock with a novel SubcorticalBlock, whose architecture draws from computational models in neuroscience and is parameterized to maximize alignment with subcortical responses reported across multiple experimental studies. Without being optimized to do so, the assembly of the SubcorticalBlock with the VOneBlock improved V1 alignment across most standard V1 benchmarks, and  better modeled extra-classical receptive field phenomena. In addition, EVNets exhibit stronger emergent shape bias and outperform the base CNN architecture by 9.3\% on an aggregate benchmark of robustness evaluations, including adversarial perturbations, common corruptions, and domain shifts. Finally, we show that EVNets can be further improved when paired with a state-of-the-art data augmentation technique, surpassing the performance of the isolated data augmentation approach by 6.2\% on our robustness benchmark. This result reveals complementary benefits between changes in architecture to better mimic biology and training-based machine learning approaches. \footnote{Code and model weights available at \url{https://github.com/lucaspiper99/evnet/}.}
\end{abstract}

\section{Introduction} \label{sec:intro}

Convolutional neural networks (CNNs) have achieved remarkable performance across a range of object recognition benchmarks \cite{NIPS2012_c399862d, simonyan2015deep, szegedy2014going, he2015deep, tan2020efficientnet}, yet they remain vulnerable when faced with common corruptions~\cite{hendrycks2019benchmarking}, domain shifts~\cite{geirhos2021partial, geirhos2022imagenettrained, geirhos2020accuracy}, and adversarial perturbations~\cite{madry2019deep, goodfellow2015explaining}. These vulnerabilities not only limit deployment in real-world settings but also underscore fundamental disparities between computer vision models and primate vision~\cite{geirhos2022imagenettrained, geirhos2021partial}. In response, recent work has introduced biologically inspired models that integrate neuroscientific computations into CNN pipelines~\cite{Dapello2020.06.16.154542, cirincione2022implementing, pushpull2024,EVANS202296, 10.1007/978-3-031-44204-9_33}. A prominent example is the VOneNet family \cite{Dapello2020.06.16.154542}, which improves adversarial and corruption robustness by combining a biologically constrained front-end --- the VOneBlock --- to conventional CNN architectures. This block simulates processing in primate primary visual cortex (V1) via a fixed-weight, empirically constrained Gabor filter bank (GFB), nonlinearities reflecting responses of V1 simple and complex cells, and a neural noise generator. While VOneNets represent a key step towards neurally-aligned vision models, they abstract away the hierarchical processing in the early visual system, notably omitting subcortical circuits such as the retina and lateral geniculate nucleus (LGN). This raises the question of whether refining upstream processing to V1, by explicitly modeling subcortical processing, yields further gains in robustness and alignment with biology. To address this question, we present the following key contributions:
\begin{itemize}
    \item We introduce a novel fixed-weight CNN front-end called the SubcorticalBlock designed to capture key computations in the retina and the LGN. This module is instantiated from neuroscientific models and is explicitly parameterized to produce responses aligned with a broad set of experimentally observed subcortical response properties.
    \item We introduce Early Vision Networks (EVNets), a new class of hybrid CNNs that combines two biologically-grounded modules, the VOneBlock with the new SubcorticalBlock, as a multi-stage front-end for a standard CNN architecture.
    \item We show that without any explicit optimization for V1 predictivity, EVNets improve neural and behavioral alignment with primate vision, outperforming both standard CNNs and VOneNets across multiple benchmarks. In particular, EVNets capture extra-classical RF phenomena more accurately, increase V1 tuning property alignment, and better mimic human inductive biases through a stronger emergence of shape bias.
    \item We demonstrate that EVNets deliver enhanced robustness across a diverse battery of evaluations, including adversarial attacks, common image corruptions, and domain shifts, and that these gains generalize to other architecture back-ends.
    \item We show that EVNets trained with a state-of-the-art (SOTA) data augmentation technique yield additive improvements in robustness, highlighting the complementary effects of architectural priors and training-based strategies.
\end{itemize}

\subsection{Related Work} \label{sec:rel_work}

\paragraph{Modeling subcortical vision.}
The Difference-of-Gaussian (DoG) model emerged as the foundational linear framework for characterizing spatial summation over the receptive field (RF) of subcortical cells \cite{kuffler1953, Rodieck1965-yv}. Subsequent extensions modeled extra-classical RF properties, including contrast gain control and surround suppression \cite{Shapley1978-dx, Solomon2006-zt}. Building on this, divisive normalization \cite{Carandini2011-xb, Bonin2005-cu} and cascading linear-nonlinear (LN) models \cite{MANTE2008625} improved subcortical predictivity by incorporating the interaction between different visual processing stages. More recently, CNNs outperformed prior models in predicting subcortical responses to visual stimuli \cite{McIntosh2016}.

\paragraph{Applications of subcortical vision.}
Beyond modeling subcortical vision, DoG-based filtering emerged as a solution to edge detection~\cite{MarrHildreth1980}, while Retinex theory~\cite{Land71} and lightness models~\cite{HORN1974277} used spatial normalization to achieve color constancy. Subsequent frameworks such as scale-space representations~\cite{Koenderink1987} and multiscale Laplacian pyramids~\cite{1095851} generalized these computations into hierarchical contrast and boundary encoding. Recent work has reintroduced these ideas into deep architectures, by embedding explicit center–surround pathways for illumination-robust classification~\cite{pmlr-v139-babaiee21a} and by unrolling Retinex-inspired optimization within CNNs for low-light image enhancement~\cite{liu2021ruas}. % Reviewer 1 asked to reference early computational models of center-surround mechanisms and DoG-based edge and contrast processing (e.g., Marr & Hildreth)

\paragraph{Improving perturbation robustness.}
CNN robustness improvements have largely been driven by data augmentation techniques~\cite{hendrycks2020augmix, hendrycks2021faces, 10.1007/978-3-031-19806-9_36}. To this end, standard benchmarks evaluate models under image corruptions~\cite{hendrycks2019benchmarking} and alternative renditions~\cite{hendrycks2021faces, salvador2022imagenetcartoon, wang2019learning, geirhos2021partial, geirhos2022imagenettrained}.  Recent work highlights that composing augmentations~\cite{hendrycks2020augmix, hendrycks2021faces,10.1007/978-3-030-58580-8_4}, especially when integrated with architectural changes~\cite{pushpull2024}, forwards the SOTA under this regime. Notably, PRIME augmentation~\cite{10.1007/978-3-031-19806-9_36} samples semantically-aligned transformations from maximum entropy distributions. In parallel, the vulnerability of CNNs to white-box adversarial perturbations has catalyzed extensive research~\cite{madry2019deep, obfuscated-gradients, carlini2019evaluating} with adversarial training emerging as the dominant paradigm for improving robustness~\cite{madry2019deep}.

\paragraph{Measuring alignment with primate vision.}
A growing suite of metrics has emerged to quantify the alignment between models and the primate vision~\cite{Marques2021.03.01.433495, Freeman2013, SchrimpfKubilius2018BrainScore, Schrimpf2020integrative}. Metrics such as shape bias~\cite{geirhos2022imagenettrained} have been instrumental in measuring model-human behavioral alignment, while a parallel line of research emphasizes representational alignment through the comparison of model tuning properties to those observed in neural data~\cite{Marques2021.03.01.433495, Freeman2013, olaiya2022measuring}. Complementing these efforts, the BrainScore platform~\cite{SchrimpfKubilius2018BrainScore, Schrimpf2020integrative} provides a unified benchmark that integrates neural recordings and behavioral data across multiple visual cortical areas, including V1, V2, V4, and IT, alongside behavioral and task-driven metrics.

\paragraph{Building neuro-inspired models.}
Introducing biological computations into CNNs has consistently enhanced their robustness to input perturbations~\cite{EVANS202296, 10.1007/978-3-031-44204-9_33, pushpull2024, 10.1007/s00521-020-04751-8, Dapello2020.06.16.154542}. Gains have been observed with convolutional layers aligned to early visual RFs~\cite{EVANS202296, 10.1007/978-3-031-44204-9_33}, push-pull inhibition motifs inspired by V1~\cite{pushpull2024, 10.1007/s00521-020-04751-8}, and by introducing the VOneBlock~\cite{Dapello2020.06.16.154542, cirincione2022implementing}.  Besides improving adversarial and corruption robustness, VOneNets achieved improved accuracy in V1 predictivity, and, when combined with V1 divisive normalization, further sharpened alignment with V1 response properties and corruption robustness \cite{cirincione2022implementing, olaiya2022measuring}. Finally, the systematic composition of hallmark V1 computations into CNNs recently achieved SOTA performance on explaining V1 predictivity and tuning properties~\cite{pogoncheff2023explaining}.

\section{Methods} \label{sec:methods}

\begin{figure}[t!]
\centering
\begin{tikzpicture}
\pgftext{\includegraphics[width=1.0\linewidth]{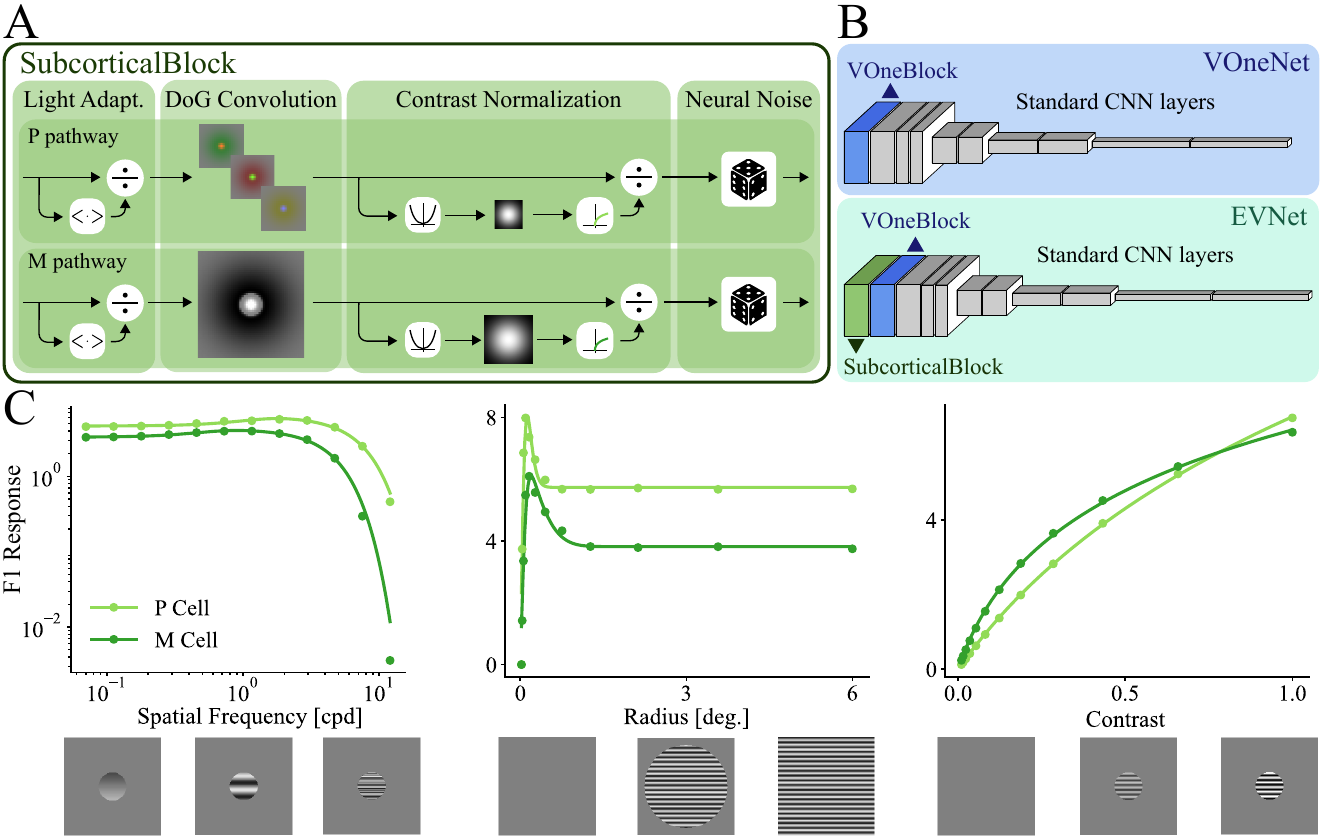}};
\end{tikzpicture}
\caption{\textbf{Simulating primate early visual processing as CNN front-end blocks.} \textbf{A} The SubcorticalBlock integrates two parallel processing pathways for P and M cells with a light-adaptation layer, a DoG convolutional layer, a contrast-normalization layer and a neural noise generator. \textbf{B} Both VOneNets and EVNets comprise an initial block designed to simulate a specific stage of the visual system, followed by a standard CNN architecture. VOneNets include a VOneBlock and EVNets include both a SubcorticalBlock and a VOneBlock. \textbf{C} SF, size, and contrast tuning curves (left to right) for two example subcortical neurons with example frames from the drifting gratings stimulus set  shown below. Markers indicate the F1 component of the cell response and the  solid line depicts the fitted response functions used for parameterizing response properties (cf. Supplementary Material~\ref{sec:sup_subcorticalblock_implementation}). Notably, the SubcorticalBlock exhibits hallmark LGN phenomena, including contrast saturation and surround suppression, with stronger modulation observed in M cells.}\label{fig:1}
\end{figure}

Inspired by the hierarchical organization of early visual processing culminating in V1, we introduce EVNets, a new family of neuro-inspired CNNs that build upon the VOneNet framework. EVNets incorporate modular fixed-weight front-ends that reflects the functional stages of the early primate visual pathway (Fig.~\ref{fig:1}). This architecture comprises three key components: the SubcorticalBlock, modeling response characteristics of foveal neurons in the retina and the LGN; a variant of the VOneBlock, which models classical RF properties of V1 neurons; and a standard CNN back-end architecture. Together, the front-end blocks instantiate a composite cascading LN model of early cortical vision. EVNets adopt a spike-count formulation, abstracting over temporal dynamics, focusing solely on spatial encoding. The EVNet  operates over a 7deg field-of-view (FoV), reduced from the 8deg used in the original VOneBlock~\cite{Dapello2020.06.16.154542}. We also extend the GFB to higher spatial frequencies (SFs), improving alignment with the SF tuning properties of primate V1~\cite{DEVALOIS1982545} (cf. Supplementary Material~\ref{sec:sup_vonenets} for an overview of VOneNets and our modifications). We trained three random seeds for both the VOneNet and the EVNet model families using a ResNet50~\cite{he2015deep} back-end architecture. All models were trained on the ImageNet-1k dataset~\cite{NIPS2012_c399862d}, with clean accuracy evaluated on the standard validation split. Additional training details are provided in Supplementary Material~\ref{sec:sup_training}.

\subsection{Architecture}

The SubcorticalBlock simulates spatial summation over the RF of parvocellular (P) and magnocellular (M) cells, processing them as separate parallel pathways. To account for both classical and extra-classical properties, each pathway comprises a light adaptation stage, a DoG convolutional layer, a contrast normalization stage, and a noise generator that simulates subcortical noise statistics.

\paragraph{DoG convolution.} Spatial summation over the RF and center-surround antagonism is modeled by incorporating a set of DoG filters described by

\begin{equation}
    \mathbf{w}_{\text{DoG}}(x,y) = \exp{\bigg(-\frac{x^2+y^2}{r_c^2}\bigg)}-(k_s/k_c)\exp{\bigg(-\frac{x^2+y^2}{r_s^2}\bigg)}\text{,}
    \label{eq:dog}
\end{equation}

\noindent where \(r_c\) and \(r_s\) are the center and surround radii, and \(k_s/k_c\) is the peak contrast sensitivity ratio. We simulate biological color-opponent pathways characteristic of the different types of cells, with the P-cell stream incorporating red-green, green-red, and blue-yellow opponency, whereas M cells reflect achromaticity by incorporating a DoG filter with no color tuning \cite{10.3389/fneur.2021.661938, Wiesel1966-zq}.

\paragraph{Light adaptation.} To mimic the subcortical mechanism of light adaptation \cite{Carandini2011-xb, MANTE2008625, berardino2018eigendistortions}, we introduce a biologically inspired module that performs global luminance normalization. This transformation modulates the input \( \mathbf{x} \) as

\begin{equation}
    \mathbf{x}_\text{LA} = \frac{\mathbf{x} - \bar{\mathbf{x}}}{\bar{\mathbf{x}}}\text{,}
    \label{eq:la}
\end{equation}

\noindent where \(\mathbf{x}_\text{LA}\) is the light-adapted input and the \( \bar{\mathbf{x}} \) denotes the average pixel intensity across channels and spatial dimensions of the input, ensuring a null output for pixel values matching the global mean.

\paragraph{Contrast normalization.} To model the adaptive gain control mechanisms characteristic of early visual processing \cite{Carandini2011-xb, Bonin2005-cu, MANTE2008625, RaghavanENEURO.0515-22.2023, Mante2005-kc}, we introduce a contrast normalization stage that normalizes activations by a local estimate of stimulus contrast. The normalized response is computed by 

\begin{equation}
    \mathbf{x_{\text{CN}}} = \frac{\mathbf{x_{\text{DoG}}}}{\big(c_{50} + \sqrt{\mathbf{x_{\text{DoG}}}^2 * \mathbf{w}_{\text{CN}}}\big)^n}\text{,}
    \label{eq:cn}
\end{equation}

\noindent where \(\mathbf{x_{\text{DoG}}}\) denotes the pre-normalized activation, \(\mathbf{w_{\text{CN}}}\) is a Gaussian kernel defining the contrast integration pooling window, \(c_{50}\) is a semi-saturation constant controlling sensitivity, and \(n\) governs the strength of the nonlinearity.

\iffalse
\noindent\begin{minipage}{.5\linewidth}
\begin{equation}
  \mathbf{x_{\text{LA}}} = \frac{\mathbf{x}^\gamma}{\mathbf{x}^\gamma + \bar{\mathbf{x}}^\gamma} - b_{\text{LA}}
    \label{eq:la}
\end{equation}
\end{minipage}%
\begin{minipage}{.5\linewidth}
\begin{equation}
  \mathbf{x_{\text{CN}}} = \frac{\mathbf{x_{\text{DoG}}}}{\big(c_{50} + \sqrt{\mathbf{x_{\text{DoG}}}^2 * \mathbf{w}_{\text{CN}}}\big)^n}
    \label{eq:cn}
\end{equation}
\end{minipage}
\fi

\paragraph{Push-pull pattern.} The canonical push-pull pattern, emerging from the antagonistic interaction between ON- and OFF-center cells in early visual circuits, can be functionally approximated by subtracting rectified responses of opposite polarity pathways \cite{doi:10.1126/science.8191289, Hirsch1998-lq}. Assuming ON and OFF cells of the same type share identical spatial profiles and gain, their RFs differ only in polarity. Under this assumption, since Equation~\ref{eq:cn} is antisymmetric with respect to \(\mathbf{x}_\text{DoG}\), the contrast-normalized responses of the ON and OFF pathways, 
\(\mathbf{x^\pm_{\text{CN}}}\), would satisfy \(\mathbf{x^+_{\text{CN}}}=-\mathbf{x^-_{\text{CN}}}\). Thus, applying rectification to both signals and computing their differences gives

\begin{equation}
    \text{max}\big(\mathbf{x^+_{\text{CN}}}, 0\big) - \text{max}\big(-\mathbf{x^+_{\text{CN}}}, 0\big) = \mathbf{x^+_{\text{CN}}}\text{.}
    \label{eq:pp2}
\end{equation}

\noindent Accordingly, our implementation bypasses explicit rectification and subtraction steps, instead operating directly on the signed contrast-normalized signal to improve computational efficiency without sacrificing functional fidelity.

\paragraph{Noise Generator.} Neuronal responses in the primate visual system exhibit trial-to-trial variability with distinct stochastic signatures across processing stages. In V1, spike count variability closely follows a Poisson distribution, where the variance is equal to the mean, corresponding to a Fano Factor equal to one~\cite{Softky1993-te}. Conversely, subcortical neurons exhibit sub-Poisson variability, characterized by a spike count variance lower than the mean~\cite{doi:10.1152/jn.00058.2022, doi:10.1152/jn.01171.2003}. To faithfully capture this hierarchical structure of neural noise, we implement a dual-source noise injection mechanism in which we add independent Gaussian noise to each unit of both front-end blocks scaled accordingly. This noise is calibrated at the unit level to maintain an overall variability with a unit Fano factor, while trial-to-trial activations at the VOneBlock output exhibit heteroskedasticity consistent with V1 measurements~\cite{Softky1993-te, }. Prior to noise injection, unit activations are linearly rescaled such that their mean response to a stimulus aligns with the empirically observed spike count of the corresponding primate neuronal population over a 50-ms integration window~\cite{doi:10.1152/jn.00058.2022, Dapello2020.06.16.154542}.

\iffalse
\begin{equation}
    {N}(\mu=0,\sigma^2=|x|F)\text{.}
    \label{eq:noise}
\end{equation}
\fi

\subsection{Subcortical-Aligning Parameterization} \label{sec:parameterization}

Despite the wealth of empirical data on primate LGN and the existence of various fitted models~\cite{Carandini2011-xb, Bonin2005-cu, MANTE2008625, RaghavanENEURO.0515-22.2023, Mante2005-kc}, the heterogeneity of modeling approaches across studies limits the direct reuse of parameters in the SubcorticalBlock, while complicating synthesis from the broader literature. To address this, we introduce a novel neurophysiologically-constrained hyperparameter tuning strategy designed to produce responses that best match the mean neuronal response properties of an LGN neuronal population taken from prior studies and modeling strategies. Specifically, we selected a total of \(N=6\) different response property distributions measured at foveal LGN to ensure alignment in SF tuning, size tuning, and contrast sensitivity. The individual properties are: center, surround, excitation and inhibition radii~\cite{Solomon338, Rodieck1965-yv, Sceniak1999}, suppression index~\cite{Solomon338} and saturation index~\cite{RaghavanENEURO.0515-22.2023}.

We conducted a series of \textit{in silico} experiments, presenting each cell with drifting gratings, quantifying each response property through the first harmonic (F1) of the cell's response. We then performed hyperparameter search via Bayesian optimization~\cite{10.1007/3-540-07165-2_55}, minimizing the loss

\begin{equation}
    \mathcal{L} = \sum_{i=1}^N \Bigg[\log_2{\bigg(\frac{R_i(\mathbf{f}_{1,i})}{\bar{r_i}}\bigg)}\Bigg]^2\text{.}
    \label{eq:loss1}
\end{equation}

\noindent For each response property \(i\), \(\bar{r}_i \) is the mean of the empirical response property distribution, \(\mathbf{f}_{1,i}\) is a vector of F1 responses produced by the SubcorticalBlock cell when subjected to the experiment-specific stimulus set, and \(R_i\) maps the F1 responses to the response property. This mapping often involved an intermediate model-fitting step, with the response property computed from the fitted parameters. Figure~\ref{fig:1}C shows the F1 responses of both P and M cells, example frames of the stimulus set used for each experiment and response model curves obtained at convergence (cf. Supplementary Material~\ref{sec:sup_subcorticalblock_implementation} for in-depth description of experiment methodology).

\subsection{Model Evaluation}

\paragraph{Alignment with primate vision.}
Shape bias was measured using a cue-conflict dataset~\cite{geirhos2022imagenettrained} that combines shapes and textures of ImageNet samples using style transfer~\cite{7780634}. To quantify the correspondence between model internal representations and V1 responses, we assessed the activations from the first block of each model using two complementary BrainScore~\cite{SchrimpfKubilius2018BrainScore, Schrimpf2020integrative} benchmarks: V1 neural predictivity~\cite{Freeman2013} and V1 response property~\cite{Marques2021.03.01.433495}. V1 neural predictivity evaluates the degree to which model features can account for the variance in primate V1 responses via partial least squares (PLS) regression. In contrast, V1 response property quantifies RF tuning similarity by comparing the distributions of 22 response properties, extracted from the same first-block activations, to empirical V1 distributions. These properties span 7 functional categories: orientation and SF tuning, response selectivity, RF size, surround modulation, texture modulation, and response magnitude.

\paragraph{Robustness evaluation.} To quantify robustness, we report a mean Robustness Score, defined as the mean top-1 accuracy across a diverse set of common corruptions, adversarial attacks, and domain shifts. To assess whether EVNets offer complementary gains to SOTA robustness training methods, we trained a standard ResNet50, a VOneResNet50 and an EVResNet50 using PRIME~\cite{10.1007/978-3-031-19806-9_36}. We included adversarial training with an \(L_\infty\) constraint of \(\|\delta\|_\infty=4/255\)~\cite{robustness} (AT\(_{L_\infty}\)) as a baseline.

\paragraph{Image corruptions.}
We evaluated model corruption robustness by measuring top-1 accuracy on the ImageNet-C dataset~\cite{hendrycks2019benchmarking}, which comprises 75 distinct corrupted variants of the ImageNet validation set. These corruptions are organized into 15 types applied at five severity levels, reflecting a specific real-world image degradation. The corruption types are further grouped into four broad categories: noise, blur, weather, and digital perturbations.

\paragraph{Domain shifts.}
To assess each model’s generalization under domain shifts, we averaged top-1 accuracies across five ImageNet-derived datasets that focus on renditions partially addressable by early visual processing mechanisms. Specifically, we used ImageNet-R~\cite{hendrycks2021faces}, ImageNet-Cartoon~\cite{salvador2022imagenetcartoon}, ImageNet-Drawing~\cite{salvador2022imagenetcartoon}, ImageNet-Sketch~\cite{wang2019learning}, and the 16-class Stylized-ImageNet~\cite{geirhos2021partial, geirhos2022imagenettrained} (Stylized\(_{16}\)-ImageNet). Each dataset introduces representational changes such as abstraction, stylization, or domain-specific distortions while preserving the underlying semantic structure.

\paragraph{Adversarial attacks.}
Following Dapello et al.~\cite{Dapello2020.06.16.154542}, we evaluated robustness to white-box adversarial attacks by applying untargeted Projected Gradient Descent (PGD)~\cite{madry2019deep} on 5000 images of the ImageNet validation set. Attacks were carried out under \(L_\infty\), \(L_2\) and \(L_1\) norm constraints and the perturbation budgets used were \(\|\delta\|_\infty\in[1/1020, 1/255, 4/255, 16/255]\), \(\|\delta\|_2\in[0.15, 0.6, 1.2, 2.4]\) and \(\|\delta\|_1\in[40, 160, 640, 2560]\). We used 64 PGD iterations with step size of \(\|\delta\|_p/32\). Additional implementation details are provided in Supplementary Material~\ref{sec:sup_adv_attacks}. %Given the stochasticity of our models, we employed the reparameterization trick~\cite{Kingma2014} and to account for the noise injection, we also performed Monte Carlo sampling to accurately estimate gradients under noise~\cite{obfuscated-gradients}, averaging 10 samples per PGD iteration. %Additional control experiments are provided in Supplementary Material~\ref{sec:sup_adv_attacks}. where we verify attack efficacy by confirming a monotonic degradation in accuracy with increasing perturbation strength, and by validating that both the number of gradient samples and the number of PGD iterations are near convergence~\cite{carlini2019evaluating}.

\paragraph{EVNet variants.} We trained seven EVResNet50 variants derived from the full EVNet by performing six targeted ablations and one architectural addition (two seeds each). The ablations individually removed the P- and M-cell pathways, the contrast normalization and light adaptation layers, the VOneBlock, and the subcortical noise generator. When removing subcortical noise, cortical noise in the VOneBlock output was amplified to remain Poisson-distributed, consistent with the original VOneNet~\cite{Dapello2020.06.16.154542}. The final variant introduced an LGN–V2 skip connection~\cite{Bullier1983} by concatenating the SubcorticalBlock output with the VOneNet bottleneck. While all EVResNet50 variants were tested for primate vision alignment and robustness, adversarial evaluations were restricted to a reduced attack set including only the two weakest perturbation strengths of each norm constraint.

\paragraph{Additional experiments.} To test whether improvements in adversarial robustness were not due to gradient masking~\cite{obfuscated-gradients}, we performed a battery of controls according to the best practices~\cite{obfuscated-gradients, Kingma2014, carlini2019evaluating} (cf. Supplementary Material~\ref{sec:sup_adv_attacks}). We further evaluated the generalization of EVNet front-ends across back-end architectures by integrating them with EfficientNet-B0~\cite{tan2020efficientnet} and CORnet-Z~\cite{23} (cf. Supplementary Material~\ref{sec:sup_backend}). Finally, we assessed performance gains from multi-pass ensemble inference in Supplementary Material~\ref{sec:ensemble_inference}.

\section{Results} \label{sec:results}

\subsection{EVNets Improve Neuronal and Behavioral Alignment with Primate Vision}

\begin{figure}[h!]
\centering
\begin{tikzpicture}
\pgftext{\includegraphics[width=1.0\linewidth]{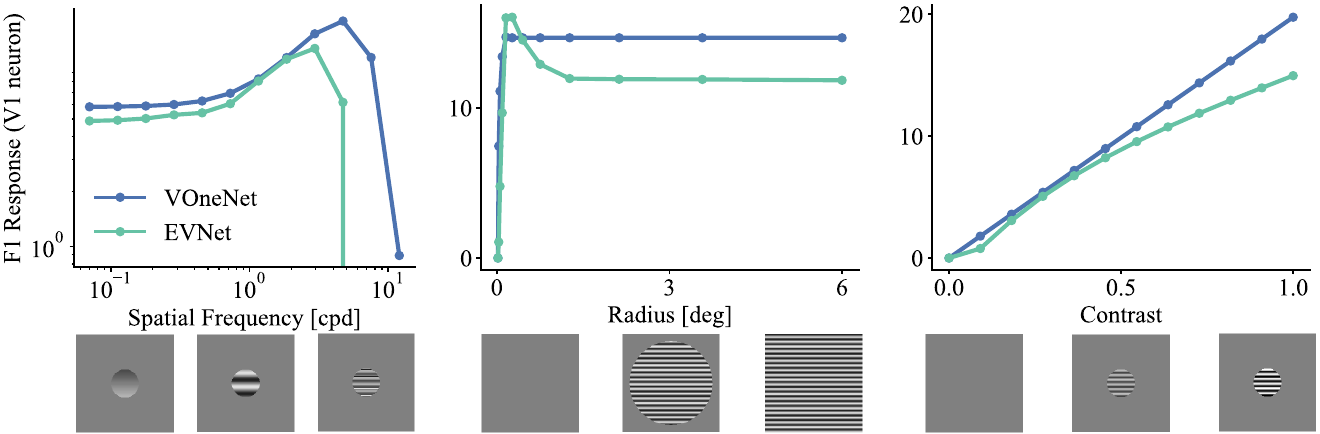}};
\end{tikzpicture}
\caption{\textbf{Subcortical preprocessing improves explanability of extra-classical RF properties in V1.} SF, size, and contrast tuning curves (left to right) for an example neuron in the VOneBlock with and without subcortical preprocessing (M cell). Example frames from the drifting gratings stimuli are shown below. VOneBlock neurons in isolation exhibit predominantly classical RF effects but when coupled with subcortical processing exhibit behaviors consistent with those empirically observed, such as enhanced surround modulation and non-linear contrast responses\cite{doi:10.1152/jn.00692.2001, SCLAR19901}. See Supplementary Material~\ref{sec:empirical_tuning_curves} for empirical V1 tuning curves.}\label{fig:2}
\end{figure}

We evaluated whether coupling the SubcorticalBlock with the VOneBlock improved alignment with primate V1. As illustrated in Figure~\ref{fig:2}, the inclusion of the SubcorticalBlock upstream of the VOneBlock introduces hallmark extra-classical RF response properties absent from the VOneBlock alone. In particular, we observe an increased surround suppression in the size tuning curve and a non-linear contrast-sensitivity curve.

Motivated by these observations, we used the BrainScore platform \cite{SchrimpfKubilius2018BrainScore, Schrimpf2020integrative} to quantitatively evaluate V1 alignment. As shown in Table~\ref{tab:brainscore}, while VOneNets outperform in V1 predictivity, EVResNet50 achieves a higher mean response property score than both the ResNet50 and VOneResNet50 models. Notably, the highest gains are observed for the surround modulation and RF size tuning, both associated with the increased surround suppression. A tradeoff in SF tuning is observed, which can be attributed to the fact that no changes were done to the GFB to account for the upstream processing.

\begin{table}[h]
\centering
\caption{\textbf{EVResNet50 outperforms baselines on mean V1 response property alignment, and shape bias.} BrainScore~\cite{SchrimpfKubilius2018BrainScore, Schrimpf2020integrative} V1 alignment scores and shape bias~\cite{geirhos2022imagenettrained} for ResNet50, VOneResNet50, and EVResNet50. Values indicate mean \(\pm\) SD (\(n=3\) seeds).}
\label{tab:brainscore}
\scalebox{0.8}{
\begin{tabular}{lcccccccccc}
\toprule
 & & & \multicolumn{7}{c}{V1 Response Properties} & \\ \cmidrule{4-10} 
 & V1 & V1 Resp. & Orient. & SF & RF & Surround & Texture & Resp. & Resp. & Shape\\
Model & Predict. & Prop. & Tuning & Tuning & Size & Mod. & Mod. & Select. & Magn. & Bias [\%]\\ \midrule
ResNet50 & .271 & .637 & .822 & .754 & .214 & .389 & .792 & .621 & .865 & 18.8\\
 & \(^{\pm\text{.002}}\) & \(^{\pm\text{.008}}\) & \(^{\pm\text{.027}}\) & \(^{\pm\text{.026}}\) & \(^{\pm\text{.002}}\) & \(^{\pm\text{.023}}\) & \(^{\pm\text{.028}}\) & \(^{\pm\text{.010}}\) & \(^{\pm\text{.012}}\) & \(^{\pm\text{1.2}}\)\\
VOneResNet50 & \textbf{.375} & .754 & \textbf{.859} & \textbf{.969} & .482 & .373 & \textbf{.919} & \textbf{.792} & .884 & 31.6\\
 & \(^{\pm\text{.002}}\) & \(^{\pm\text{.006}}\) & \(^{\pm\textbf{.005}}\) & \(^{\pm\textbf{.001}}\) & \(^{\pm\text{.041}}\) & \(^{\pm\text{.003}}\) & \(^{\pm\textbf{.004}}\) & \(^{\pm\textbf{.003}}\) & \(^{\pm\text{.002}}\) & \(^{\pm\text{1.2}}\)\\ 
\textbf{EVResNet50} & .364 & \textbf{.826} & .854 & .950 & \textbf{.726} & \textbf{.614} & .916 & .781 & \textbf{.933} & \textbf{48.9}\\
 & \(^{\pm\text{.000}}\) & \(^{\pm\textbf{.000}}\) & \(^{\pm\text{.009}}\) & \(^{\pm\text{.000}}\) & \(^{\pm\textbf{.001}}\) & \(^{\pm\textbf{.004}}\) & \(^{\pm\text{.001}}\) & \(^{\pm\text{.000}}\) & \(^{\pm\textbf{.000}}\) & \(^{\pm\textbf{2.4}}\)\\ \bottomrule
\end{tabular}}
\end{table}

Among the suit of primate visual behavior alignment metrics, shape bias has emerged as a particularly informative proxy of human-aligned inductive biases and out-of-domain (OOD) generalization~\cite{geirhos2022imagenettrained, geirhos2021partial}. Motivated by the hypothesis that shape bias may originate in early visual computations, we evaluated shape bias in EVNets (Tab.~\ref{tab:brainscore}) and observed a substantial increase of 30.1\% relative to the standard ResNet50 and of 17.3\% relative to the VOneNet model, suggesting that EVNets may confer not only improved neuronal alignment but also behavioral traits more consistent with primate perception.

\subsection{EVNets Improve Robustness Across an Aggregate Benchmark}

When tested on clean images (Tab.~\ref{tab:corruptions}), our VOneNets variant achieves a 1.2\% improvement over the original VOneNet~\cite{Dapello2020.06.16.154542}, while EVNets displayed a performance drop of 1.3\%, when compared to the same reported accuracy. We further examined whether the increased alignment of EVNets with primate vision, also leads to improved robustness.

\paragraph{Image corruptions.} Across most corruption categories and in terms of mean corruption accuracy, EVNets consistently outperformed both VOneNets and the base ResNet50, which only retained an advantage in weather corruptions. Notably, the most pronounced gains were observed on noise corruptions, where EVNets outperformed VOneNets by 3.7\% while effectively preserving the same cumulative Fano factor as VOneNets (cf. Fig.~\ref{fig:all_corr} for accuracy across individual corruptions).

\begin{table}[t]
\caption{\textbf{EVResNet50 outperforms baselines on most image corruption types and on mean corruption accuracy.} Clean and corrupted top-1 accuracies averaged across severities and corruptions for ResNet50, VOneResNet50 and EVResNet50. Values indicate mean \(\pm\) SD (\(n=3\) seeds).}
\label{tab:corruptions}
\centering
\begin{tabular}{lcccccc}
\toprule
 & & \multicolumn{4}{c}{Corruption Types} & \\ \cmidrule(lr){3-6}
 & Mean & Noise & Blur & Weather & Digital & Clean \\
 Model & {[}\%{]} & {[}\%{]} & {[}\%{]} & {[}\%{]} & {[}\%{]} & {[}\%{]} \\ \midrule
ResNet50 & 38.8\scriptsize{\(\pm\)0.5} & 29.2\scriptsize{\(\pm\)0.6} & 34.6\scriptsize{\(\pm\)0.4} & \textbf{36.1}\scriptsize{\(\boldsymbol{\pm}\)\textbf{0.5}} & 49.5\scriptsize{\(\pm\)0.6} & \textbf{75.4}\scriptsize{\(\boldsymbol{\pm}\)\textbf{0.1}} \\
VOneResNet50 & 40.4\scriptsize{\(\pm\)0.1} & 35.9\scriptsize{\(\pm\)0.5} & 34.8\scriptsize{\(\pm\)0.1} & 32.6\scriptsize{\(\pm\)0.1} & 52.2\scriptsize{\(\pm\)0.1} & 72.9\scriptsize{\(\pm\)0.1} \\
\textbf{EVResNet50}  & \textbf{41.9}\scriptsize{\(\boldsymbol{\pm}\)\textbf{0.2}} & \textbf{39.6}\scriptsize{\(\boldsymbol{\pm}\)\textbf{0.3}} & \textbf{37.5}\scriptsize{\(\boldsymbol{\pm}\)\textbf{0.1}} & 30.6\scriptsize{\(\pm\)0.2} & \textbf{53.5}\scriptsize{\(\boldsymbol{\pm}\)\textbf{0.1}} & 70.4\scriptsize{\(\pm\)0.1}\\ \bottomrule
\end{tabular}
\end{table}

\paragraph{Domain shifts.}
Table~\ref{tab:ood} summarizes our OOD generalization results. EVNets consistently outperforms both the baseline ResNet50 and VOneNets across the majority of benchmarks, also surpassing these baselines on the mean domain shift accuracy. The only dataset where EVNets underperform is ImageNet-Sketch, where the base ResNet50 exhibits a marginal advantage.

\begin{table}[t]
\centering
\caption{\textbf{EVResNet50 outperforms baselines on most domain shift datasets and on overall mean domain shift accuracy.} Top-1 accuracies on ImageNet-\{Cartoon, Drawing, R, Sketch, Stylized\(_{16}\)\} for ResNet50, VOneResNet50 and EVResNet50. Values indicate mean \(\pm\) SD (\(n=3\) seeds).}
\label{tab:ood}
\begin{tabular}{lcccccc}
\toprule
 & Mean & Cartoon & Drawing & R & Sketch & Stylized\(_{16}\) \\
Model & {[}\%{]} & {[}\%{]} & {[}\%{]} & {[}\%{]} & {[}\%{]} & {[}\%{]} \\ \midrule
ResNet50 & 33.4\scriptsize{\(\pm\)0.2} & 51.2\scriptsize{\(\pm\)0.7} & 20.9\scriptsize{\(\pm\)0.6} & 35.4\scriptsize{\(\pm\)0.1} & \textbf{23.3}\scriptsize{\(\boldsymbol{\pm}\)\textbf{0.1}} & 36.3\scriptsize{\(\pm\)1.2} \\
VOneResNet50 & 37.1\scriptsize{\(\pm\)0.4} & 55.5\scriptsize{\(\pm\)0.2} & 30.5\scriptsize{\(\pm\)0.4} & 37.5\scriptsize{\(\pm\)0.1} & 23.1\scriptsize{\(\pm\)0.3} & \textbf{38.8}\scriptsize{\(\boldsymbol{\pm}\)\textbf{1.1}} \\
\textbf{EVResNet50} & \textbf{38.1}\scriptsize{\(\boldsymbol{\pm}\)\textbf{0.3}} & \textbf{57.1}\scriptsize{\(\boldsymbol{\pm}\)\textbf{0.3}} & \textbf{33.9}\scriptsize{\(\boldsymbol{\pm}\)\textbf{0.2}} & \textbf{38.1}\scriptsize{\(\boldsymbol{\pm}\)\textbf{0.2}} & 22.7\scriptsize{\(\pm\)0.2} & 38.6\scriptsize{\(\pm\)1.1} \\ \bottomrule
\end{tabular}
\end{table}

\paragraph{Adversarial attacks.}
EVNets improves adversarial robustness across most perturbation norms and attack strengths when compared to VOneNets (Tab.~\ref{tab:adversarial}). While VOneNets obtained a marginal advantage under the weakest attack strengths for \(L_\infty\) and \(L_2\) norm constraints, EVNets improved robustness on the remaining attack settings, exhibiting also a smaller gap between clean and adversarial accuracy.

\begin{table}[h]
\centering
\caption{\textbf{EVResNet50 outperforms baselines on most adversarial perturbations and on mean adversarial robustness.} Top-1 accuracies for the ResNet50, VOneResNet50 and EVResNet50 models. Values indicate mean \(\pm\) SD (\(n=3\) seeds).}
\label{tab:adversarial}
\scalebox{0.85}{
\begin{tabular}{lccccccccccccc}
\toprule
 & & \multicolumn{4}{c}{\(\|\delta\|_\infty\)} & \multicolumn{4}{c}{\(\|\delta\|_2\)} & \multicolumn{4}{c}{\(\|\delta\|_1\)} \\ \cmidrule(lr){3-6} \cmidrule(lr){7-10} \cmidrule(lr){11-14}
 & Mean & \(\frac{1}{1020}\) & \(\frac{1}{255}\) & \(\frac{4}{255}\) & \(\frac{16}{255}\) & 0.15 & 0.6 & 2.4 & 9.6 & 40 & 160 & 640 & 2560 \\
Model & {[}\%{]} & {[}\%{]} & {[}\%{]} & {[}\%{]} & {[}\%{]} & {[}\%{]} & {[}\%{]} & {[}\%{]} & {[}\%{]} & {[}\%{]} & {[}\%{]} & {[}\%{]} & {[}\%{]} \\ \midrule
ResNet50 & 8.3 & 23.4 & 0.4 & 0.2 & \textbf{0.2} & 37.2 & 1.8 & 0.2 & \textbf{0.2} & 33.6 & 1.7 & 0.2 & \textbf{0.2} \\
& \(^{\pm{0.2}}\) & \(^{\pm{0.8}}\) & \(^{\pm{0.0}}\) & \(^{\pm{0.0}}\) & \(^{\boldsymbol{\pm}\textbf{0.0}}\) & \(^{\pm{1.0}}\) & \(^{\pm{0.2}}\) & \(^{\pm{0.0}}\) & \(^{\boldsymbol{\pm}\textbf{0.0}}\) & \(^{\pm{0.7}}\) & \(^{\pm{0.2}}\) & \(^{\pm{0.0}}\) & \(^{\boldsymbol{\pm}\textbf{0.0}}\) \\
VOneResNet50 & 26.1 & 62.6 & 30.4 & 1.2 & 0.0 & \textbf{66.2} & 42.3 & 4.2 & 0.0 & \textbf{64.5} & 37.3 & 3.9 & 0.0 \\
& \(^{\boldsymbol{\pm}\textbf{0.1}}\) & \(^{\pm{0.4}}\) & \(^{\pm{0.3}}\) & \(^{\pm{0.1}}\) & \(^{\pm{0.0}}\) & \(^{\boldsymbol{\pm}\textbf{0.3}}\) & \(^{\pm{0.2}}\) & \(^{\pm{0.1}}\) & \(^{\pm{0.0}}\) & \(^{\boldsymbol{\pm}\textbf{0.3}}\) & \(^{\pm{0.6}}\) & \(^{\pm{0.2}}\) & \(^{\pm{0.0}}\) \\
\textbf{EVResNet50} & \textbf{28.3} & \textbf{62.7} & \textbf{38.8} & \textbf{3.0} & 0.0 & 65.1 & \textbf{48.0} & \textbf{7.4} & 0.0 & 64.0 & \textbf{44.5} & \textbf{6.0} & 0.0 \\
& \(^{\boldsymbol{\pm}\textbf{0.2}}\) & \(^{\boldsymbol{\pm}\textbf{0.2}}\) & \(^{\boldsymbol{\pm}\textbf{0.6}}\) & \(^{\boldsymbol{\pm}\textbf{0.1}}\) & \(^{\pm{0.0}}\) & \(^{\pm{0.0}}\) & \(^{\boldsymbol{\pm}\textbf{0.3}}\) & \(^{\boldsymbol{\pm}\textbf{1.8}}\) & \(^{\pm{0.0}}\) & \(^{\pm{0.2}}\) & \(^{\pm\textbf{0.4}}\) & \(^{\pm\textbf{1.8}}\) & \(^{\pm{0.0}}\) \\ \bottomrule
\end{tabular}}
\end{table}

Table~\ref{tab:robust} summarizes all the robustness results described above and presents the Robustness Score for the evaluated models. The EVResNet50 model improves the Robustness Score by 9.3\% over the base ResNet50 and by 1.6\% over our VOneNet variant.

\subsection{Combining EVNets with Data Augmentation Provides Cumulative Gains}

In our evaluation of the Robustness Score for the EVResNet50 with PRIME data augmentation (Tab.~\ref{tab:robust}), we find that this combined strategy yields cumulative performance gains beyond those of either component alone. Furthermore, when comparing these results with those obtained by augmenting a ResNet50 with adversarial training and augmenting a VOneNet with PRIME, no configuration surpasses the additive gains achieved by the EVResNet50 trained with PRIME.

\begin{table}[h]
\centering
\caption{\textbf{EVResNet50 achieves a higher Robust Score than ResNet50 and VOneResNet50 and, when coupled with PRIME, surpasses SOTA data augmentation approaches.} Robustness Score, clean and perturbed top-1 accuracies for the ResNet50, VOneResNet50 and EVResNet50 models; for ResNet50 with two data augmentation approaches: AT\(_{L_\infty}\) and PRIME; and for VOneResNet50 and EVResNet50 with PRIME. Values indicate mean \(\pm\) SD (\(n=3\) seeds).}
\label{tab:robust}
\begin{tabular}{lccccc}
\toprule
 &  & \multicolumn{3}{c}{Perturbations} & \\ \cmidrule{3-5}
 & Robust. Score & Adversarial & Corrupt. & Domain Shift & Clean \\
Model & {[}\%{]} & {[}\%{]} & {[}\%{]} & {[}\%{]} & {[}\%{]} \\ \midrule
ResNet50 & 26.8\scriptsize{\(\pm\)0.2} & 8.3\scriptsize{\(\pm\)0.2} & 39.1\scriptsize{\(\pm\)0.3} & 33.3\scriptsize{\(\pm\)0.1} & \textbf{75.4}\scriptsize{\(\boldsymbol{\pm}\)\textbf{0.1}} \\
VOneResNet & 34.5\scriptsize{\(\pm\)0.1} & 26.1\scriptsize{\(\pm\)0.1} & 40.4\scriptsize{\(\pm\)0.2} & 37.1\scriptsize{\(\pm\)0.4} & 72.9\scriptsize{\(\pm\)0.0} \\
\textbf{EVResNet50} & \textbf{36.1}\scriptsize{\(\boldsymbol{\pm}\)\textbf{0.2}} & \textbf{28.3}\scriptsize{\(\boldsymbol{\pm}\)\textbf{0.2}} & \textbf{41.9}\scriptsize{\(\boldsymbol{\pm}\)\textbf{0.2}} & \textbf{38.1}\scriptsize{\(\boldsymbol{\pm}\)\textbf{0.3}} & 70.4\scriptsize{\(\pm\)0.1} \\ \midrule
ResNet50 + AT\(_{L_\infty}\)~\cite{robustness} & 34.4 & \textbf{31.3} & 32.5 & 39.5 & 62.4 \\
ResNet50 + PRIME & 36.5\scriptsize{\(\pm\)0.1} & 14.3\scriptsize{\(\pm\)0.2} & 52.6\scriptsize{\(\pm\)0.2} & 42.8\scriptsize{\(\pm\)0.2} & \textbf{76.0}\scriptsize{\(\boldsymbol{\pm}\)\textbf{0.1}} \\
VOneResNet50 + PRIME & 42.1\scriptsize{\(\pm\)0.1} & 28.7\scriptsize{\(\pm\)0.3} & \textbf{53.4}\scriptsize{\(\boldsymbol{\pm}\)\textbf{0.0}} & \textbf{44.3}\scriptsize{\(\boldsymbol{\pm}\)\textbf{0.3}} & 74.0\scriptsize{\(\pm\)0.1} \\
\textbf{EVResNet50} + PRIME & \textbf{42.7}\scriptsize{\(\boldsymbol{\pm}\)\textbf{0.2}} & 30.7\scriptsize{\(\pm\)0.8} & 53.2\scriptsize{\(\pm\)0.1} & 44.2\scriptsize{\(\pm\)0.1} & 72.0\scriptsize{\(\pm\)0.1} \\ \bottomrule
\end{tabular}
\end{table}

\subsection{EVNet Variants Reveal Competing Drivers of Primate Vision Alignment and Robustness}

Selective ablation of specific SubcorticalBlock components affected primate vision alignment in different aspects (see Supplementary Material~\ref{sec:sup_variant}). The components that caused a greater drop in V1 response property benchmarks were the M-cell pathway and the contrast normalization, which greatly affected the RF size and surround modulation. Surprisingly, removing either the P-cell pathway or light adaptation greatly increased shape bias. 

\begin{table}[h]
\centering
\caption{\textbf{Model components contribution to robustness vary greatly.} Robustness Score, clean and perturbed top-1 accuracies for all EVResNet50 variants. ResNet50 and VOneResNet50 included for reference but not in the comparison. Values indicate mean \(\pm\) SD (\(n=2\) seeds).}
\label{tab:variants_robust}
\begin{tabular}{lccccc}
\toprule
 &  & \multicolumn{3}{c}{Perturbations} & \\ \cmidrule{3-5}
 & Robust. Score\(^\ast\) & Adversarial\(^\ast\) & Corrupt. & Domain Shift & Clean \\
Model & {[}\%{]} & {[}\%{]} & {[}\%{]} & {[}\%{]} & {[}\%{]} \\ \midrule
\textcolor{gray}{ResNet50} & \textcolor{gray}{29.5\scriptsize{\(\pm\)0.3}} & \textcolor{gray}{16.4\scriptsize{\(\pm\)0.5}} & \textcolor{gray}{39.1\scriptsize{\(\pm\)0.3}} & \textcolor{gray}{33.3\scriptsize{\(\pm\)0.1}} & \textcolor{gray}{75.4\scriptsize{\(\pm\)0.1}} \\
\textcolor{gray}{VOneResNet50} & \textcolor{gray}{42.7\scriptsize{\(\pm\)0.1}} & \textcolor{gray}{50.5\scriptsize{\(\pm\)0.1}} & \textcolor{gray}{40.4\scriptsize{\(\pm\)0.2}} & \textcolor{gray}{37.1\scriptsize{\(\pm\)0.4}} & \textcolor{gray}{72.9\scriptsize{\(\pm\)0.0}} \\
EVResNet50 & 44.6\scriptsize{\(\pm\)0.1} & 53.8\scriptsize{\(\pm\)0.2} & 41.9\scriptsize{\(\pm\)0.2} & 38.1\scriptsize{\(\pm\)0.3} & 70.5\scriptsize{\(\pm\)0.0} \\
\(-\) P Cells & 38.4\scriptsize{\(\pm\)0.0} & 46.7\scriptsize{\(\pm\)0.2} & 34.9\scriptsize{\(\pm\)0.1} & 33.6\scriptsize{\(\pm\)0.0} & 60.7\scriptsize{\(\pm\)0.1} \\
\(-\) M Cells & \textbf{44.8\scriptsize{\(\pm\)0.1}} & 54.0\scriptsize{\(\pm\)0.2} & 42.2\scriptsize{\(\pm\)0.1} & 38.2\scriptsize{\(\pm\)0.3} & 70.3\scriptsize{\(\pm\)0.1} \\
\(-\) Light Adapt. & 44.3\scriptsize{\(\pm\)0.3} & \textbf{55.1\scriptsize{\(\pm\)0.5}} & 40.4\scriptsize{\(\pm\)0.2} & 37.4\scriptsize{\(\pm\)0.2} & 69.8\scriptsize{\(\pm\)0.2} \\
\(-\) Contrast Norm. & 44.4\scriptsize{\(\pm\)0.2} & 53.5\scriptsize{\(\pm\)0.1} & 41.7\scriptsize{\(\pm\)0.4} & 38.0\scriptsize{\(\pm\)0.1} & 70.7\scriptsize{\(\pm\)0.0} \\
\(-\) Subcort. Noise & 42.6\scriptsize{\(\pm\)0.0} & 48.5\scriptsize{\(\pm\)0.1} & 41.9\scriptsize{\(\pm\)0.2} & 37.4\scriptsize{\(\pm\)0.1} & \textbf{72.6\scriptsize{\(\pm\)0.2}} \\
\(-\) VOneBlock & 44.3\scriptsize{\(\pm\)0.3} & 51.8\scriptsize{\(\pm\)0.5} & \textbf{42.7\scriptsize{\(\pm\)0.2}} & 38.2\scriptsize{\(\pm\)0.1} & 71.6\scriptsize{\(\pm\)0.0} \\
\(+\) LGN-V2 Connect. & \textbf{44.8\scriptsize{\(\pm\)0.0}} & 53.9\scriptsize{\(\pm\)0.2} & 42.1\scriptsize{\(\pm\)0.1} & \textbf{38.3\scriptsize{\(\pm\)0.2}} & 70.7\scriptsize{\(\pm\)0.1} \\ \bottomrule
\end{tabular}

\footnotesize{\(^\ast\) Computed with a reduced attack set (two perturbations per norm constraint); not comparable to Table~\ref{tab:robust}.}
\end{table}

In terms of performance, the single component that had the largest impact on both clean accuracy and robustness was the P-cell pathway (Tab.~\ref{tab:variants_robust}). On the other hand, removing the M-cell pathway or contrast normalization had no impact on clean accuracy and robustness. Light adaptation had only minor effects on specific robustness benchmarks and a small drop in clean accuracy. Removing subcortical stochasticity improved clean accuracy while decreasing adversarial robustness.  Interestingly, the omission of the VOneBlock barely affected overall robustness, revealing even to be the best variant under image corruptions and an improvement in clean accuracy, accompanied by a small drop in adversarial robustness. The inclusion of LGN-V2 skip connections had little to no impact in performance across all perturbations types and clean images.

\section{Discussion} \label{sec:discussion}

In this work, we present a new family of neuro-inspired CNNs that not only display enhanced robustness across a broad spectrum of perturbations but also achieve stronger alignment with primate vision. Previous works have incorporated biologically inspired mechanisms such as DoG filtering~\cite{pmlr-v139-babaiee21a, 10.1007/978-3-031-44204-9_33, EVANS202296}, divisive normalization~\cite{cirincione2022implementing, miller2022divisive}, and noise injection~\cite{10.1007/978-3-030-58580-8_4} within CNN architectures. In contrast, our SubcorticalBlock introduces a principled segregation of computations into parallel P- and M-cell pathways constrained by prior empirical neurophysiological findings. % Reviewer 1 asked to mention that P and M-cell pathways are key source of novelty (unlike prior DoG filtering, divisive normalization, and noise injection strategies)
The EVNet architecture introduces a modular, cascading model of V1 that incorporates stage-specific architectural priors, offering a compelling alternative to V1-inspired CNN paradigms. The V1-alignment improvements observed in EVNets are most pronounced for surround modulation and RF size. These effects likely arise from the normalization mechanisms within the SubcorticalBlock, which induce local competition and lead to extraclassical RF responses not present in VOneNets. Despite these improvements, benchmark scores for surround modulation and RF size are still relatively low, suggesting that normalization mechanisms at the V1 level, either caused by recurrent or feedback circuits, are also needed for a better alignment~\cite{cirincione2022implementing, miller2022divisive}. % Reviewer 3 asked to discuss how subcortical computations influence V1 tuning properties
While EVNets do not entirely solve the longstanding problem of robust generalization, our results also underscore the critical importance of subcortical processing in shaping early visual representations: while the DoG filtering improves features selectivity, performs low-pass filtering, mitigating high-frequency noise, the normalization layers promote local competition and dynamic range compression, reducing sensitivity to input perturbations. % Reviewer 3 asked to discuss source of robustness gains
Notably, we demonstrate that meaningful improvements to V1 modeling can be achieved by exclusively refining upstream stages. The cumulative gain shown by combining the SubcorticalBlock with the VOneBlock reveals a biologically-plausible potential for compositionality in inductive biases, where each module targets distinct axes of visual invariance, together producing synergistic improvements in perturbation robustness. Specifically, while the SubcorticalBlock primarily encodes invariance to luminance and contrast, the VOneBlock focuses more on spatial and polarity invariance. Finally, we show that integrating biologically-inspired architectures with standard data augmentation techniques leads to synergistic improvements in robustness, surpassing data augmentation alone and adversarial training, the most effective methods for improving adversarial and corruption robustness. % Reviewer 4 asked to better contextualize our findings within broader computer vision research.
Together, these results provide further evidence that neuroscience-driven inductive biases and machine learning heuristics are not mutually exclusive, but can in fact be complementary.

While the observed robustness gains are compelling, there are also some trade-offs. Specifically, our model abstracts subcortical processing by instantiating only four channels that reflect the average spatial response profiles observed in the LGN, rather than capturing the full heterogeneity of subcortical cells. An interesting observation is that the M-cell pathway contributes marginally to downstream performance, which is in line with the classical view of M cells small contribution to the ventral stream. However, this result presents some inconsistencies with more recent literature~\cite{CANARIO2016110}. Additionally, enhancements in perturbation robustness are consistently accompanied by modest reductions in clean image accuracy, a common tension in robustness research~\cite{pushpull2024, Dapello2020.06.16.154542}. 
Beyond these considerations, subsequent work could examine the possibility of initializing from neuro-inspired weights while allowing task-driven fine-tuning of the front-ends used. % Reviewer 4 asked to discuss the possibility of initializing from neuro-inspired weights but allowing task-driven fine-tuning
For V1 alignment, we kept the VOneBlock unchanged as upstream processing minimally affected its responses aside from the improved extra-classical RF effects. However, the slight drop in V1 alignment on some benchmarks likely reflects unforseen interactions between modules. Future work could adapt the GFB to account for subcortical preprocessing and add normalization mechanisms at the V1 level to evaluate if these changes further enhance alignment and robustness. Finally, while our approach may seem to contrast the “Bitter Lesson”~\cite{sutton2019bitter}, the trajectory of neurally aligned vision tells a more nuanced story. Prior work indicates that scaling alone does not yield more brain-like representations~\cite{NEURIPS2019_7813d159}, and large-scale benchmarks~\cite{gokce2025scaling} show that neural alignment with primate visual cortex saturates with model size --- architectures with stronger biological priors often align better than transformers. Our work leverages this insight by embedding early vision–inspired inductive biases to achieve the efficiency and alignment characteristic of biological vision. % Reviewer 4 asked to discuss reflection on the tension between scaling and structure

\paragraph{Broader impact.} 
This work advances computer vision by proposing a biologically grounded alternative to vision transformers~\cite{dosovitskiy2021an}. By simulating early vision, we bridge the gap between human and machine vision, improving robustness, transparency, and interpretability, with potential benefits for bias reduction and accessibility. Fully replicating human vision remains challenging and calls for further research. We acknowledge that some foundational insights stem from animal studies, emphasizing the ethical responsibility to minimize harm and favor alternative approaches. Leveraging existing biological data, as done here, may reduce reliance on new animal experimentation.

%%%%%%%%%%%%%%%%%%%%%%%%%%%%%%%%%%%%%%%%%%%%%%%%%%%%%%%%%%%%

\clearpage

\section*{Acknowledgments}

%This work was supported by the project Center for Responsible AI reference no. C628696807-00454142, financed by the Recovery and Resilience Facility, by project PRELUNA, grant PIDC/CCIINF/4703/2021, and by Fundação para a Ciência e Tecnologia (FCT), under grant UIDB/50021/2020.

This work was supported by the project Center for Responsible AI reference no. C628696807-00454142, and by national funds through Fundação para a Ciência e a Tecnologia, I.P. (FCT) under projects UID/50021/2025 and UID/PRR/50021/2025.

\printbibliography

\clearpage
\appendix \label{sec:sup}
\renewcommand\thefigure{\Alph{section}\arabic{figure}}
\renewcommand{\theHfigure}{A\arabic{figure}}
\renewcommand\thetable{\Alph{section}\arabic{table}}
\renewcommand{\theHtable}{A\arabic{figure}}
\appendixpage

\section{VOneNets} \label{sec:sup_vonenets}
\setcounter{figure}{0}
\setcounter{table}{0}

\begin{SCfigure}[][h!]
    \includegraphics[width=0.485\linewidth]{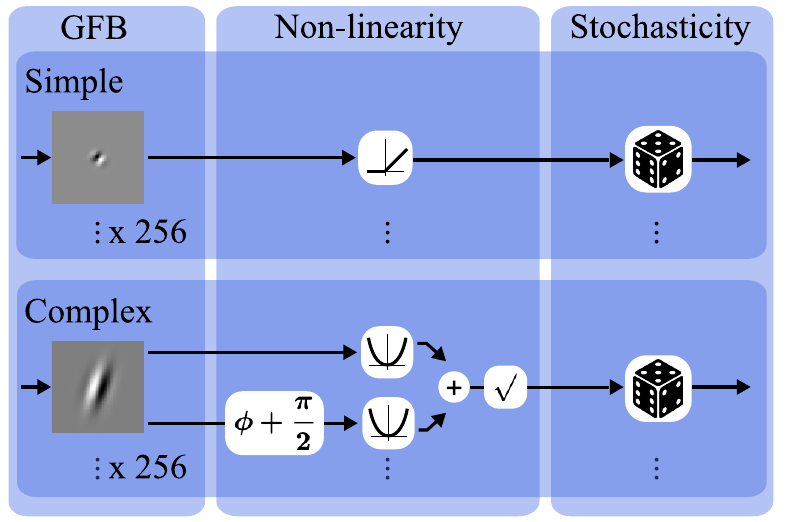}
    \caption{\textbf{VOneNets simulate V1 processing upstream of standard CNNs.} Each VOneNet incorporates a biologically-constrained front-end, the VOneBlock, preceding a conventional CNN. The VOneBlock consists of a fixed-weight Gabor filter bank (GFB) parameterized by empirical distributions, nonlinearities emulating simple and complex cell responses, and a stochastic component that injects Poisson-like noise to mimic V1 neuronal variability. Adapted from Baidya et al.~\cite{baidya2021combining}.}\label{fig:a1}
\end{SCfigure}

VOneNets \cite{Dapello2020.06.16.154542} are convolutional neural networks (CNNs) augmented with a biologically-inspired, fixed-weight front-end simulating primary visual cortex (V1), termed the VOneBlock. This front-end is structured as a linear–nonlinear–Poisson (LNP) cascade \cite{RUST2005945}, incorporating a Gabor filter bank (GFB) \cite{Jones1987-dy}, nonlinearities for both simple and complex cells \cite{Adelson:85}, and a stochastic spiking mechanism modeling neuronal variability \cite{Softky1993-te} (cf. Fig.~\ref{fig:a1}). The GFB parameters are sampled from empirical distributions of orientation preference, spatial frequency (SF) tuning, and receptive field (RF) size \cite{DEVALOIS1982545, DEVALOIS1982531, ringach2002spatial_dtructure}. The channels are split evenly between simple and complex cells, and a Poisson-like noise generator is applied to emulate spiking variability. The full implementation is available at \url{https://github.com/dicarlolab/vonenet} under a GNU General Public License v3.0.

To construct a VOneNet, we replaced the initial block of each base architecture with the VOneBlock, along with a channel-matching bottleneck layer to maintain architectural compatibility between the front-end and the downstream convolutional stack. Consistent with the original VOneResNet50 model~\cite{Dapello2020.06.16.154542}, we replaced a single convolutional layer, batch normalization, nonlinearity, and a max-pooling operation and preserved the original configuration of 512 channels within the VOneBlock, allocating 256 channels to each cell type (simple and complex). When using an EfficientNet-B0, we substituted the initial convolution, batch normalization, and activation with the VOneBlock. Since this initial block has a total stride of 2, we decreased the stride of the VOneBlock accordingly. Furthermore, given the reduced channel dimensionality in the EfficientNet-B0 where the second stage expects only 32 channels, in contrast to 64 in ResNet50 we downscaled the VOneBlock, employing 128 channels per cell type. 
Finally, for the CORnet-Z we removed a single convolutional layer, nonlinearity, and max-pooling operation. Since this first block has the same combined stride and output dimension as the ResNet50, no additional modifications were necessary.

We modified the VOneBlock by adjusting the field of view (FoV) to 7deg (down from the original 8deg) and increased the SF range of the GFB to 0.5 -- 8.0 cpd (from 0.5 -- 5.6cpd). This modification allows us to better match empirical V1 distributions, while maintaining a similar safety margin with respect to the Nyquist SF. Additionally, to maintain consistency with upstream processing, we configured the GFB to uniformly sample a single channel from the input, regardless of any preceding subcortical transformations. Finally, to ensure methodological consistency in the spike-based activation regime across both the SubcorticalBlock (cf. Section~\ref{sec:sup_subcorticalblock_implementation}) and the VOneBlock, we imposed a unified temporal integration window of 50ms. In alignment with Table C2 of Dapello et al.~\cite{Dapello2020.06.16.154542}, we applied a linear scaling factor to the VOneBlock outputs such that the mean evoked response to a batch of natural images from ImageNet matched the target stimulus response of 0.655~spikes. This scaling factor was computed independently for the VOneNet and for each EVnet variant described in this study.

\clearpage
\section{Primate Vision Alignment} \label{sec:sup_alignment}

\subsection{Empirical V1 Tuning Curves} \label{sec:empirical_tuning_curves}

\begin{figure}[h!]
\centering
\begin{tikzpicture}
\pgftext{\includegraphics[width=1\linewidth]{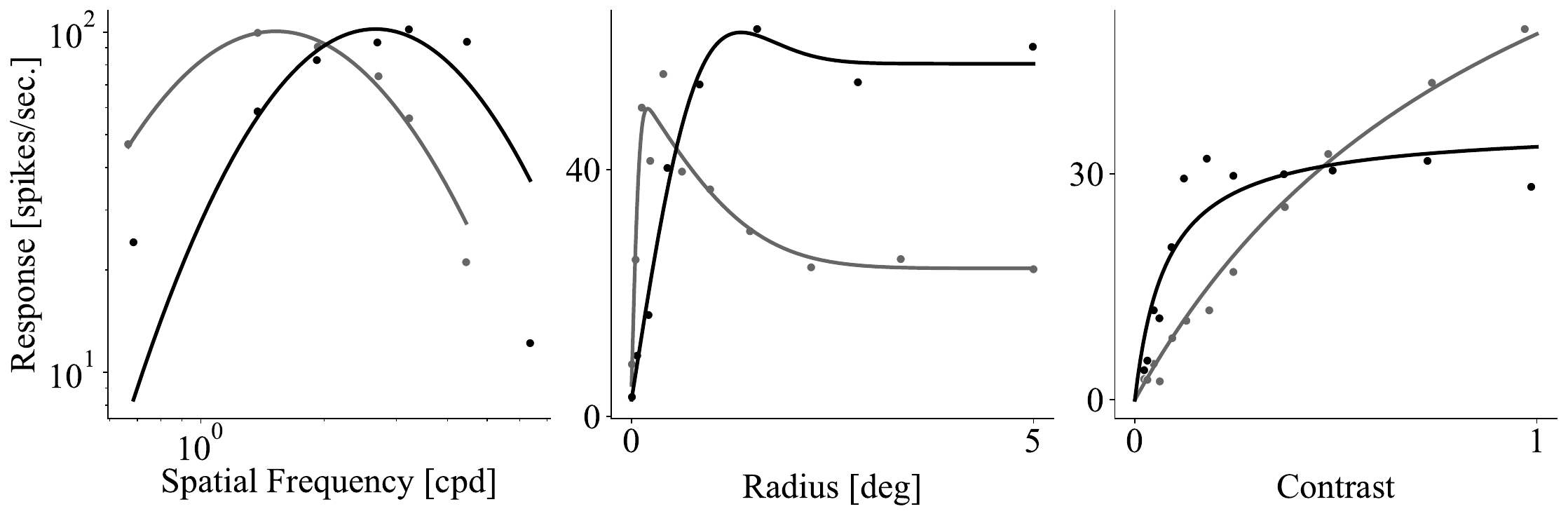}};
\end{tikzpicture}
\caption{\textbf{Examples of empirical V1 tuning curves retrieved from the literature.} SF, size, and contrast tuning curves (left to right) for example V1 neurons. \textbf{Left} SF tuning curve of a simple (gray) and complex (black) cell to drifting grating stimuli. Markers represent the total number of F1 responses to gratings of different SF normalized to the best response and the solid line depicts a quadratic fit for purposes of illustrating the tuning profile (adapted from Figure 10 in Schiller et al.~\cite{doi:10.1152/jn.1976.39.6.1334}). \textbf{Middle} Size tuning curve of two complex cells of V1 with distinct degrees of surround suppression under high contrasts. Gray depicts the a cell form V1 layer 4B under 0.15 contrast and black represents a cell from V1 layer 6 under 0.31 contrast. Markers represent each cell's F1 response to differently-sized gratings and the line depicts the predicted response of a fitted DoG model discussed in the original article (adapted from Figure 1 in Sceniak et al.~\cite{Sceniak1999}). \textbf{Right} Contrast tuning curve of two simple V1 cell from the least (gray) and most (black) contrast sensitive thirds of their respective population. Marks indicate F1 response and the solid line depicts a fitted response model discussed in the original article (adapted from Figure 2 in Sclar et al.~\cite{SCLAR19901}). Data points extracted via WebPlotDigitalizer~\cite{WebPlotDigitizer}.}\label{fig:v1_tuning_empirical}
\end{figure}

\subsection{Shape-bias}

In contrast to humans, who predominantly rely on shape cues for object recognition, ImageNet-trained CNNs have been shown to exhibit a strong bias toward texture-based representations~\cite{geirhos2022imagenettrained}. Measuring shape bias thus serves as a proxy for alignment with human inductive biases. We evaluate this using the cue conflict dataset from Geirhos et al.~\cite{geirhos2022imagenettrained}, where images contain conflicting shape and texture cues (e.g., a cat-shaped image with elephant texture). While humans tend to classify by shape, ImageNet-trained CNNs often prefer texture. A model’s shape bias is computed as the proportion of shape-consistent predictions out of all shape- or texture-consistent responses.

\subsection{BrainScore}

The BrainScore platform~\cite{SchrimpfKubilius2018BrainScore, Schrimpf2020integrative} is a standardized benchmarking suite for evaluating how brain-like artificial neural networks (ANNs) are. In the context of object recognition, BrainScore compares model activations against neural recordings from primate visual areas and human behavioral data. For early visual processing, V1 predictivity is quantified via the \texttt{FreemanZiemba2013}~\cite{Freeman2013} neural benchmark, while response properties in V1 are assessed using the \texttt{Marques2020}~\cite{Marques2021.03.01.433495} benchmark. BrainScore aggregates multiple such benchmarks into a composite score that reflects a model's alignment with neural and behavioral patterns observed in biological systems. Researchers can submit their models for evaluation at \url{https://www.brain-score.org/}.

\begin{table}[h]
\centering
\caption{\textbf{Detailed results for V1 response property alignment.} BrainScore~\cite{SchrimpfKubilius2018BrainScore, Schrimpf2020integrative} V1 alignment scores for ResNet50, VOneResNet50, and EVResNet50. Values indicate mean \(\pm\) SD (\(n=3\) seeds).}
\label{tab:resp_prop}
\begin{tabular}{llccc}
\toprule
 & & & Models & \\ \cmidrule{3-5}
Category & Resp. Property & ResNet50~\cite{he2015deep} & VOneResNet50 & EVResNet50 \\ \midrule
 Orientation & Orientation Selective & 0.975\scriptsize{\(\pm\)0.024} & \textbf{0.999\scriptsize{\(\pm\)0.001}} & \textbf{0.999\scriptsize{\(\pm\)0.001}} \\
  & Circ. Variance (CV) & \textbf{0.818\scriptsize{\(\pm\)0.011}} & 0.742\scriptsize{\(\pm\)0.013} & 0.754\scriptsize{\(\pm\)0.001} \\
  & Orth./Pref. Ratio & \textbf{0.855\scriptsize{\(\pm\)0.023}} & 0.717\scriptsize{\(\pm\)0.014} & 0.710\scriptsize{\(\pm\)0.001} \\
  & CV Bandwidth Ratio & 0.740\scriptsize{\(\pm\)0.024} & \textbf{0.763\scriptsize{\(\pm\)0.005}} & 0.762\scriptsize{\(\pm\)0.001} \\
  & Pref. Orientation & 0.943\scriptsize{\(\pm\)0.046} & \textbf{0.985\scriptsize{\(\pm\)0.004}} & 0.968\scriptsize{\(\pm\)0.000} \\
  & Orth./Pref.-CV Diff. & 0.766\scriptsize{\(\pm\)0.016} & \textbf{0.885\scriptsize{\(\pm\)0.004}} & 0.869\scriptsize{\(\pm\)0.001} \\
  & Or. Bandwidth & 0.659\scriptsize{\(\pm\)0.086} & 0.922\scriptsize{\(\pm\)0.010} & \textbf{0.952\scriptsize{\(\pm\)0.000}} \\ \midrule
 Spatial & Peak SF & 0.551\scriptsize{\(\pm\)0.047} & \textbf{0.961\scriptsize{\(\pm\)0.002}} & \textbf{0.961\scriptsize{\(\pm\)0.001}} \\
 Frequency & SF Bandwidth & 0.826\scriptsize{\(\pm\)0.019} & \textbf{0.962\scriptsize{\(\pm\)0.006}} & 0.937\scriptsize{\(\pm\)0.000} \\
  & SF Selective & 0.886\scriptsize{\(\pm\)0.053} & \textbf{0.983\scriptsize{\(\pm\)0.005}} & 0.951\scriptsize{\(\pm\)0.000} \\ \midrule
 Response & Texture Selective & 0.678\scriptsize{\(\pm\)0.008} & \textbf{0.800\scriptsize{\(\pm\)0.004}} & 0.774\scriptsize{\(\pm\)0.001} \\
 Selectivity & Modulation Ratio & 0.349\scriptsize{\(\pm\)0.009} & \textbf{0.737\scriptsize{\(\pm\)0.002}} & 0.736\scriptsize{\(\pm\)0.000} \\
  & Texture Var. Ratio & \textbf{0.794\scriptsize{\(\pm\)0.014}} & 0.703\scriptsize{\(\pm\)0.011} & 0.694\scriptsize{\(\pm\)0.001} \\
  & Texture Sparseness & 0.663\scriptsize{\(\pm\)0.032} & \textbf{0.927\scriptsize{\(\pm\)0.002}} & 0.920\scriptsize{\(\pm\)0.000} \\ \midrule
 RF Size & Grating Sum. Field & 0.272\scriptsize{\(\pm\)0.005} & 0.547\scriptsize{\(\pm\)0.016} & \textbf{0.716\scriptsize{\(\pm\)0.003}} \\
  & Surround Diameter & 0.156\scriptsize{\(\pm\)0.000} & 0.361\scriptsize{\(\pm\)0.015} & \textbf{0.736\scriptsize{\(\pm\)0.000}} \\ \midrule
 Surround Mod. & Surround Sup. Index & 0.389\scriptsize{\(\pm\)0.023} & 0.373\scriptsize{\(\pm\)0.003} & \textbf{0.614\scriptsize{\(\pm\)0.004}} \\ \midrule
 Texture & Abs. Texture Mod. Idx. & \textbf{0.978\scriptsize{\(\pm\)0.019}} & 0.942\scriptsize{\(\pm\)0.004} & 0.934\scriptsize{\(\pm\)0.000} \\
 Modulation & Texture Mod. Idx. & 0.606\scriptsize{\(\pm\)0.040} & 0.897\scriptsize{\(\pm\)0.011} & \textbf{0.898\scriptsize{\(\pm\)0.001}} \\ \midrule
 Response & Max. Texture & 0.939\scriptsize{\(\pm\)0.002} & 0.906\scriptsize{\(\pm\)0.010} & \textbf{0.951\scriptsize{\(\pm\)0.001}} \\
 Magnitude & Max. DC & \textbf{0.873\scriptsize{\(\pm\)0.053}} & 0.824\scriptsize{\(\pm\)0.008} & 0.885\scriptsize{\(\pm\)0.001} \\
  & Max. Noise & 0.783\scriptsize{\(\pm\)0.018} & 0.923\scriptsize{\(\pm\)0.006} & \textbf{0.965\scriptsize{\(\pm\)0.000}} \\
  \bottomrule
\end{tabular}
\end{table}

\subsection{V1 Predictivity}

To predict model’s ability to predict single-neuron responses in V1, we employed a dataset~\cite{Freeman2013} comprising responses from 102 V1 neurons to 450 unique 4deg image patches, spanning both naturalistic textures and noise stimuli. Predictivity was measured as the explained variance using partial least squares (PLS) regression under a 10-fold cross-validation scheme

\subsection{V1 Response Properties}

Marques et al.~\cite{Marques2021.03.01.433495} introduced a novel model-to-brain comparison framework that bypasses conventional fitting procedures, instead relying on \textit{in silico} neurophysiology to establish direct, one-to-one correspondences between artificial and V1 neurons. By probing models with canonical stimulus sets such as drifting gratings and texture pattern, the method quantifies alignment through a normalized similarity metric grounded in the Kolmogorov-Smirnov distance, capturing the distributional match of neural response properties. Critically, this framework enables rigorous benchmarking against prior neurophysiological studies without requiring raw recordings, effectively transforming existing literature into executable V1-aligning tests. In total, this method focuses on 22 distinct response characteristics, organized into seven functional domains: orientation tuning, spatial frequency tuning, receptive field size, surround modulation, texture modulation, response selectivity, and response magnitude. Table~\ref{tab:resp_prop} presents all individual response properties along with the scores obtained for the ResNet50, the VOneResNet50 and EVResNet50 models.

\clearpage
\section{Image Perturbations} \label{sec:sup_perturbations}
\setcounter{table}{0}
\setcounter{figure}{0}

\subsection{Adversarial Attacks} \label{sec:sup_adv_attacks}

To evaluate white-box robustness, we employed Projected Gradient Descent (PGD)~\cite{madry2019deep} on top of a subset of 5000 images from the ImageNet validation split. PGD is a widely adopted first-order attack that has proven effective against biologically inspired models such as VOneNets~\cite{Dapello2020.06.16.154542}. We selected this attack due to its scalability to large datasets (e.g., 5k ImageNet samples), and its compatibility with deterministic inference pipelines, which avoids the pitfalls of stochastic defenses that may artificially degrade attack success rates~\cite{Dapello2020.06.16.154542}. We run PGD for \(N=64\) iterations, where each step follows the update rule:

\[
    \mathbf{x}^{t+1} = \mathcal{P}_{\mathbf{x} + \mathcal{S}}\left( \mathbf{x}^t + \alpha \ \mathrm{sgn}\left( \nabla_{\mathbf{x}^t} \mathcal{L}(\theta, \mathbf{x}^t, \mathbf{y}) \right) \right),
\]

\noindent where \(\mathbf{x}^t\) denotes the adversarial input at iteration \(t\), \(\mathcal{L}\) is the cross-entropy loss function, and \(\mathcal{P}_{\mathbf{x} + \mathcal{S}}\) projects back onto the perturbation set \(\mathcal{S}\) centered at the clean input \(\mathbf{x}\). Under an \(L_\infty\) threat model, \(\mathcal{S}\) corresponds to a box constraint, while for \(L_1\) or \(L_2\) norms with a perturbation budget \(\epsilon\), the gradient direction is rescaled at each iteration to have the respective norm \(\alpha\), and the projection ensures the final adversarial input \(\mathbf{x}_\text{adv}\) satisfies \(\|\mathbf{x}_\text{adv} - \mathbf{x}\|_p \leq \epsilon\). We used a used a step size of \(\alpha = \epsilon / 32\) and performed a total of 12 attacks carried out under \(L_\infty\), \(L_2\) and \(L_1\) norm constraints at four perturbation budgets each: \(\|\delta\|_\infty\in[1/1020, 1/255, 4/255, 16/255]\), \(\|\delta\|_2\in[0.15, 0.6, 1.2, 2.4]\) and \(\|\delta\|_1\in[40, 160, 640, 2560]\). We used the Adversarial Robustness ToolBox v1.17.1~\cite{art2018} to conduct all the attacks.

\begin{figure}[h!]
\centering
\begin{tikzpicture}
\pgftext{\includegraphics[width=1\linewidth]{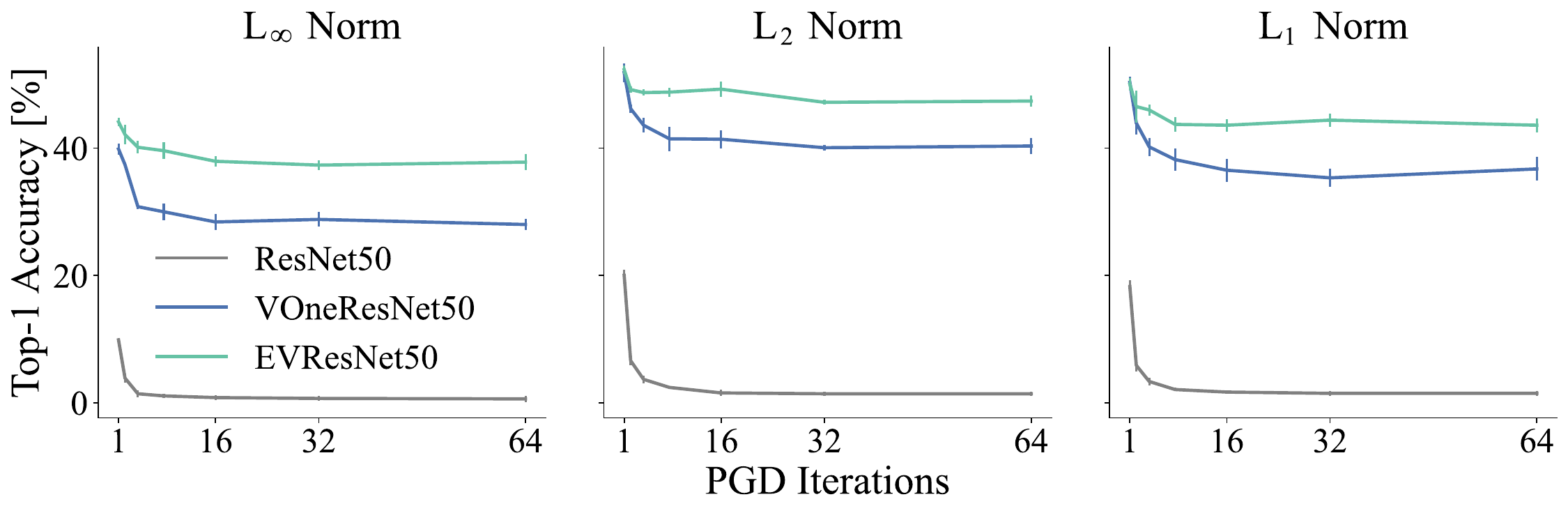}};
\end{tikzpicture}
\caption{\textbf{Adversarial robustness is evaluated at convergance of PGD iterations.} Top-1 white-box accuracy iteration curves for PGD attacks with \(\|\delta\|_\infty=1/255\), \(\|\delta\|_2=0.6\), \(\|\delta\|_1=160\) constraints for ResNet50, VOneResNet50 and EVResNet50 models, evaluated on 500 images. The step size was adjusted to be \(\epsilon\) for 1 iterations, and \(2\epsilon/N\), in the remaining cases. Increasing the number of PGD iteration steps increases attack effectiveness only up to roughly 32 iterations. Lines indicate the mean accuracy and shaded error bars denote SD (\(n=3\) seeds).}\label{fig:iterations_control}
\end{figure}

Due to the inherent stochasticity in our models, special considerations were necessary to enable gradient-based adversarial optimization. We first applied the reparameterization trick~\cite{Kingma2014}, which permits gradient flow through stochastic nodes by expressing random variables as deterministic functions of noise. To obtain reliable gradient estimates for PGD, we further adopted the approach of Athalye et al.~\cite{obfuscated-gradients}, replacing \(\nabla f\) with an average over multiple stochastic forward passes. Specifically, we estimate gradients as

\[
\nabla f \approx \frac{1}{k} \sum_{i=1}^k \nabla_i f,
\]

\noindent where each \(\nabla_i f\) corresponds to a gradient computed using an independent Monte Carlo sample. We set \(k = 10\), similarly to prior work on VOneNets, given that the additional noise source in the SubcorticalBlock did not mask the gradients any further.

\begin{SCfigure}[][h!]
    \includegraphics[width=0.485\linewidth]{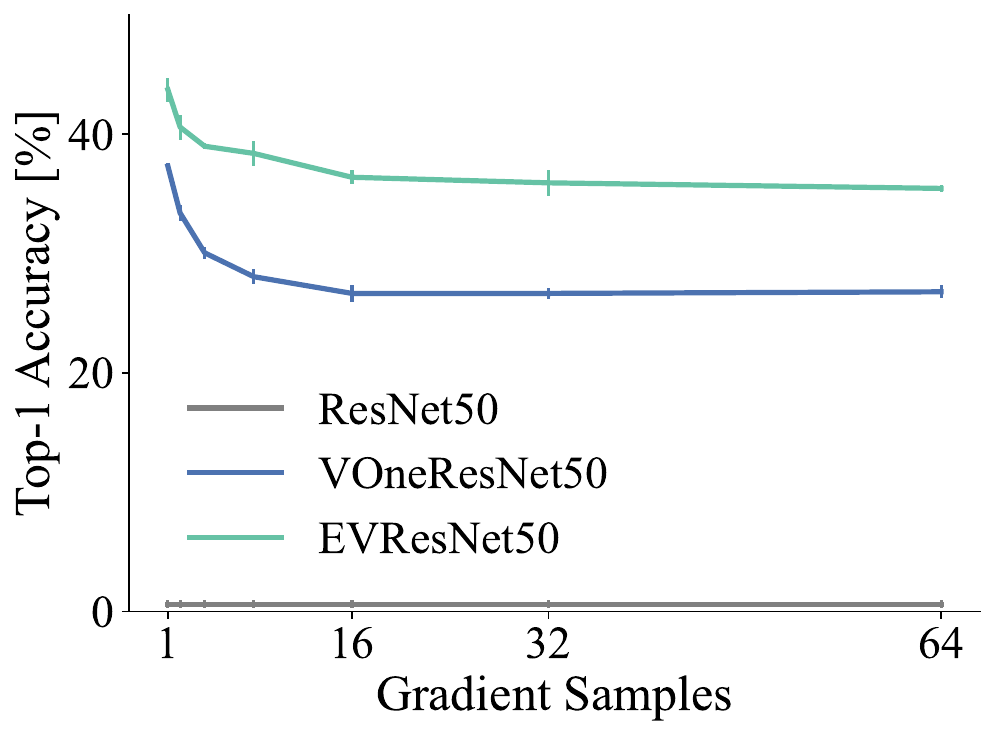}
    \caption{\textbf{Increasing the number of Monte Carlo gradient samples has limited impact on white-box attack effectiveness.} White-box PGD accuracy evaluated under \(\|\delta\|_\infty=1/255\) with 64 PGD iterations on 500 images. Increasing \(k\) from 10 to 64 leads to only marginal decreases in accuracy for both VOneResNet50 and EVResNet50, indicating that additional samples do not substantially strengthen the attack. Lines indicate the mean accuracy and error bars denote SD (\(n=3\) seeds).}\label{fig:c1}
\end{SCfigure}

To verify the reliability of our adversarial evaluation pipeline, we conducted a suite of sanity checks on a 500-image subset from the ImageNet validation set. In line with the recommendations of Athalye et al.~\cite{obfuscated-gradients} and Carlini et al.~\cite{carlini2019evaluating}, we confirmed that top-1 accuracy decreases monotonically as a function of perturbation strength across all norm constraints (Tab.~4). Additionally, we verified that increasing the number of PGD iterations increased attack effectiveness (Fig.~\ref{fig:iterations_control}) and that increasing the number of gradient samples in the Monte Carlo approximation did not lead to a substantial increase in attack success (Fig.~\ref{fig:c1}), further supporting the completeness of our threat model.

\subsection{Image Corruptions} \label{sec:sup_corruptions}

\begin{figure}[h!]
\centering
\begin{tikzpicture}
\pgftext{\includegraphics[width=.8\linewidth]{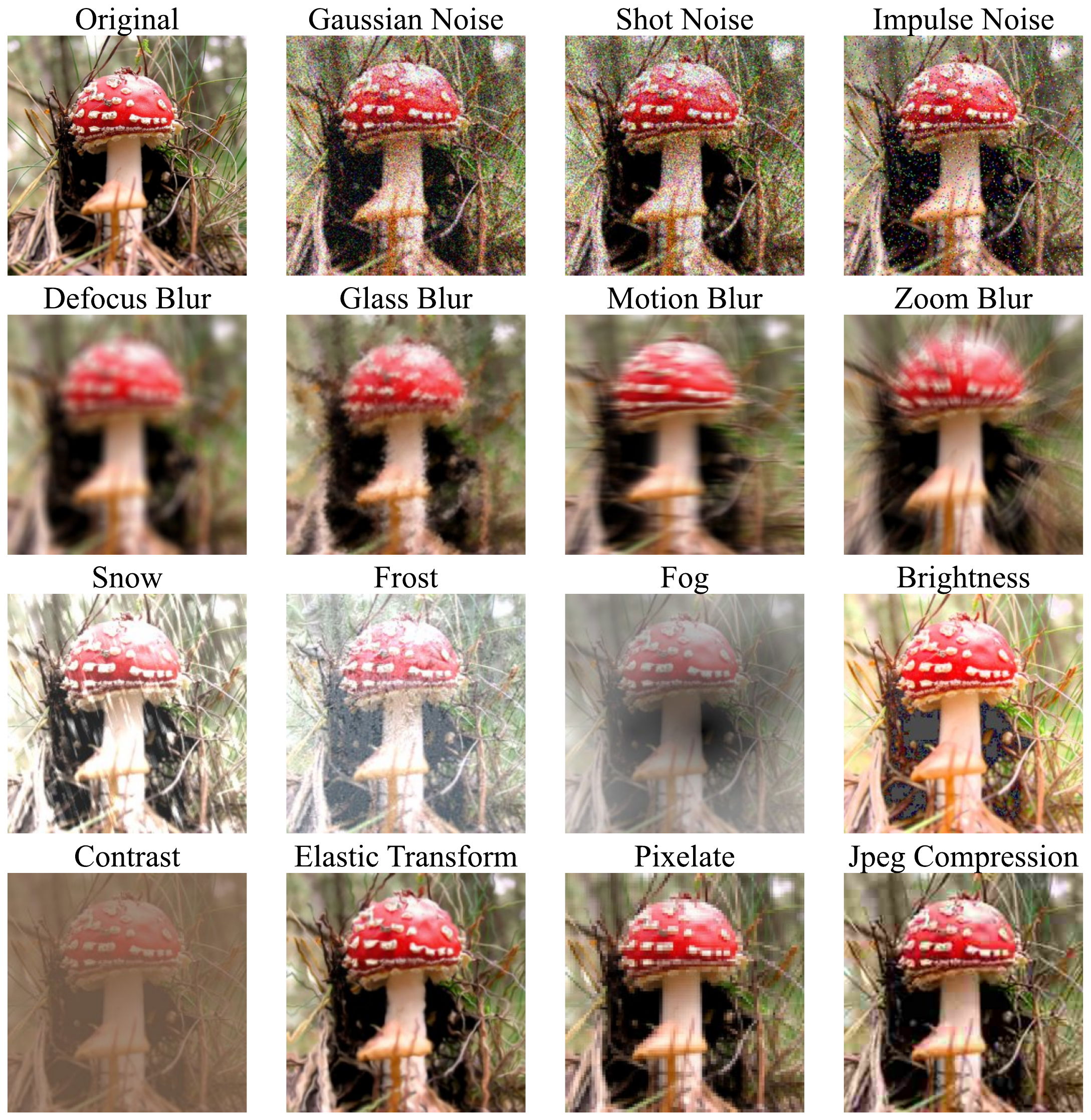}};
\end{tikzpicture}
\caption{\textbf{Examples of common image corruptions from the ImageNet-C dataset at intermediate severity (level 3)} The first row shows the original image and three noise corruptions; the second row displays blur corruptions; the third row presents weather-related corruptions; and the fourth row illustrates digital corruptions.}\label{fig:c2}
\end{figure}

%\begin{figure}[h!]
%\centering
%\begin{tikzpicture}
%\pgftext{\includegraphics[width=1\linewidth]{figc3.pdf}};
%\end{tikzpicture}
%\caption{\textbf{EVResNet50 enhances robustness to noise, blur, and digital corruptions.} Top-1 accuracy across corruption severity levels for the four categories of common corruptions: noise, blur, weather, and digital. Solid lines indicate the mean performance; error bars denote SD (\(n=3\)seeds)}\label{fig:c3}
%\end{figure}

\begin{figure}[h]
\centering
\begin{tikzpicture}
\pgftext{\includegraphics[width=.9\linewidth]{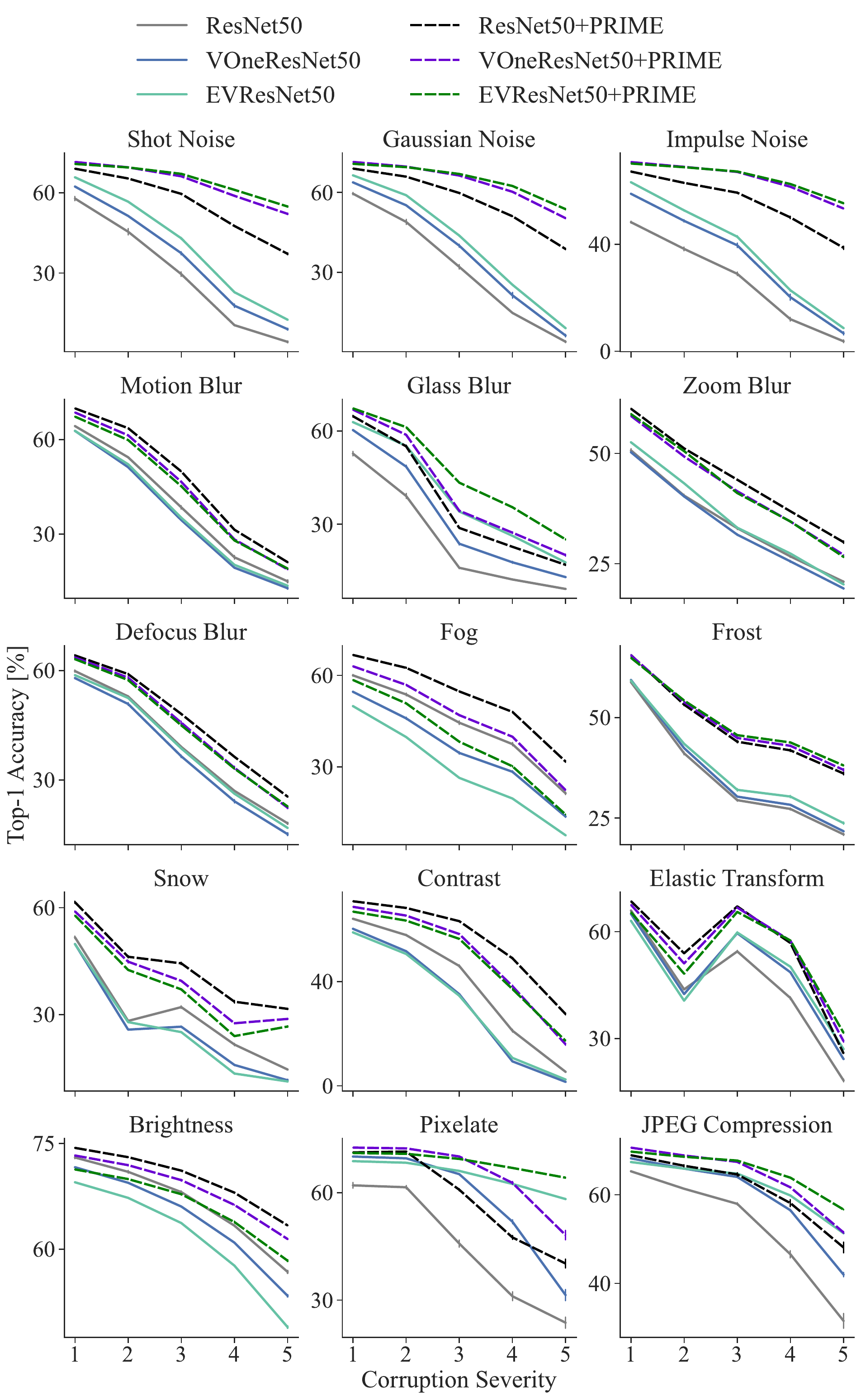}};
\end{tikzpicture}
\caption{\textbf{Detailed results for common corruptions benchmarks.} Top-1 accuracy across 5 severity levels for the 15 individual common corruptions of ImageNet-C. Lines indicate the mean top-1 accuracy and error bars denote SD (\(n=3\) seeds)}\label{fig:all_corr}
\end{figure}

The ImageNet-C dataset~\cite{hendrycks2019benchmarking} consists of 15 different corruption types, each at 5 levels of severity for a total of 75 different perturbations applied to validation images of the ImageNet validation split. Accuracy improvement on these datasets should be indicative of model robustness gains, given that it comprises, in total, 75 diverse corruptions. The individual corruption types are Gaussian noise, shot noise, impulse noise, defocus blur, glass blur, motion blur, zoom blur, snow, frost, fog, brightness, contrast, elastic transform, pixelated, and JPEG compression. The individual corruption types are grouped into 4 categories: noise, blur, weather, and digital effects. Examples of image corruptions are presented in Figure~\ref{fig:c2}. The ImageNet-C dataset is publicly available at \url{https://github.com/hendrycks/robustness} under Creative Commons Attribution 4.0 International.

\subsection{Domain Shifts} \label{sec:sup_shifts}

\paragraph{ImageNet-R}
The ImageNet-R dataset \cite{hendrycks2021faces}, consists of a curated set of 200 classes from the ImageNet validation set. This dataset includes 30,000 images featuring renditions in various artistic styles, such as paintings, sketches, and cartoons, designed to test a model's ability to generalize beyond natural image statistics. ImageNet-R is publicly available at \url{https://github.com/hendrycks/imagenet-r}.

\paragraph{ImageNet-Cartoon \& ImageNet-Drawing}
The ImageNet-Cartoon and ImageNet-Drawing datasets \cite{salvador2022imagenetcartoon}, are two domain shift benchmarks derived from the ImageNet validation set by applying label-preserving style transformations. ImageNet-Cartoon contains images transformed into cartoon-like renditions using a GAN-based framework \cite{9157493}, while ImageNet-Drawing comprises colored pencil sketch versions of the same images created via an image processing pipeline \cite{10.5555/2330147.2330161}. These datasets challenge models to generalize beyond natural image statistics, revealing significant accuracy drops—on average 18 and 45 percentage points, respectively—when standard ImageNet-trained models are evaluated. Both datasets are publicly available at \url{https://zenodo.org/records/6801109} under Creative Common Attribution 4.0 International.

\paragraph{ImageNet-Sketch.}
The ImageNet-Sketch dataset \cite{wang2019learning} is a large-scale benchmark designed to evaluate OOD generalization in image classification. It contains 50,000 black-and-white sketch-style images, with 50 images for each of the 1,000 classes in the ImageNet validation set, collected independently using keyword queries like “sketch of \texttt{[class name]}”. Unlike perturbation-based datasets, ImageNet-Sketch represents a significant domain shift in both texture and color, challenging models trained on natural images to rely on global structure rather than local textural cues. The dataset is publicly available at \url{https://www.kaggle.com/datasets/wanghaohan/imagenetsketch}.

\paragraph{Stylized\(_\text{16}\)-ImageNet.}
Stylized-ImageNet \cite{geirhos2022imagenettrained} is created by introducing different painting styles into ImageNet images through Adaptive Instance Normalization style transfer \cite{huang2017arbitrary}. While texture cues are replaced by those in the paintings, overall shape is preserved. Since the original dataset was introduced primarily for training purposes and models exhibited extremely low performance, we instead used a subset of Stylized-ImageNet as used in Geirhos et al. \cite{geirhos2021partial}. This subset focuses on 16 basic categories (e.g., airplane, dog) that are supersets of 227 ImageNet classes within the WordNet hierarchy \cite{10.1145/219717.219748}. We followed the same approach as the original article, where the probability distribution over ImageNet classes is mapped to this 16-class distribution by averaging the probabilities of corresponding fine-grained classes. This 16-class Stylized ImageNet along with the code for probability aggregation is publicly available at \url{https://github.com/bethgelab/model-vs-human}.

\begin{figure}[h!]
\centering
\begin{tikzpicture}
\pgftext{\includegraphics[width=.85\linewidth]{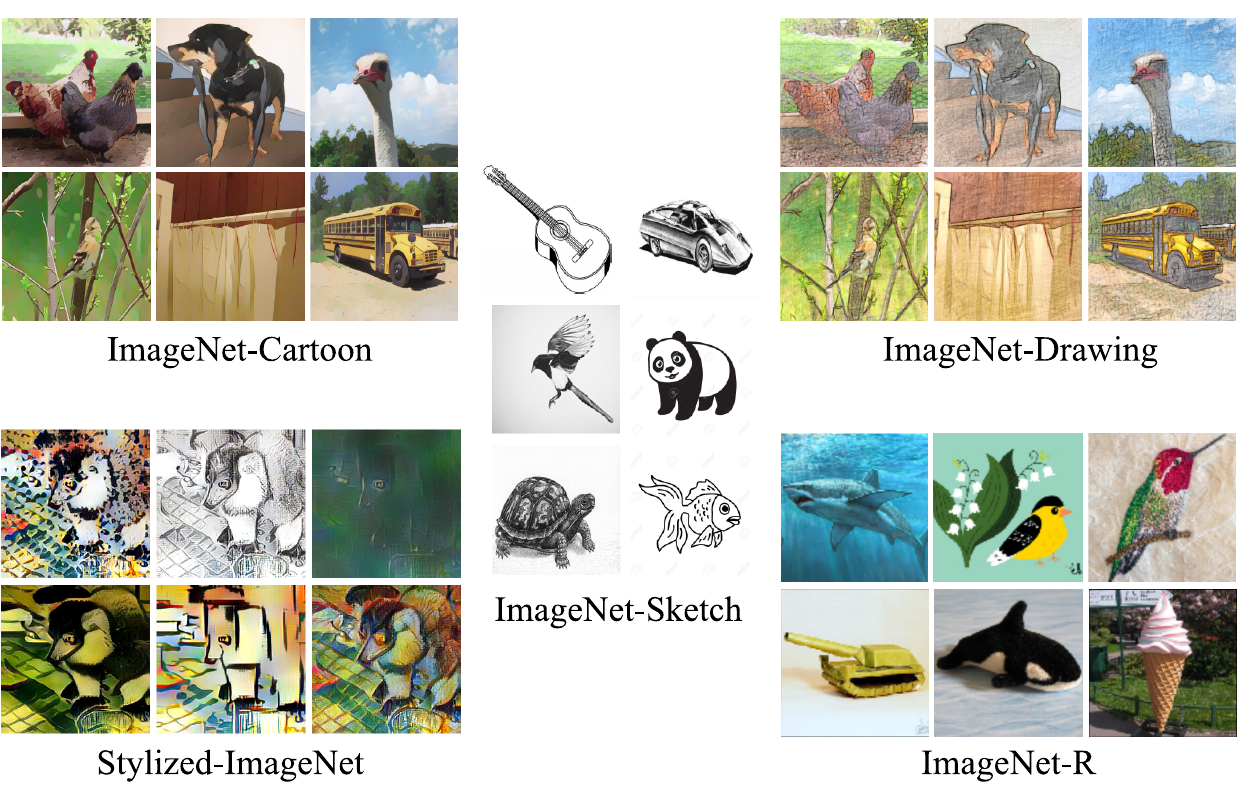}};
\end{tikzpicture}
\caption{\textbf{Examples from domain shift benchmark datasets derived from ImageNet.} \textbf{ImageNet-Cartoon} (adopted from Salvador and Oberman~\cite{salvador2022imagenetcartoon}) features object representations in cartoon style. \textbf{ImageNet-Drawing} (adopted from Salvador and Oberman~\cite{salvador2022imagenetcartoon}) includes images  rederd as colored pencil-like hand drawings. \textbf{ImageNet-Sketch} (adopted from Wang et al.~\cite{wang2019learning}) consists of black-and-white sketches emphasizing contours. \textbf{Stylized-ImageNet} (adopted from Geirhos et al.~\cite{geirhos2022imagenettrained}) applies replaces the original textures with those of random paintings. \textbf{ImageNet-R} (adopted from Henrdycks et al.~\cite{hendrycks2021faces}) contains various renditions of ImageNet classes, including painting, cartoon, origami, toy, embroidery, and sculpture styles.}\label{fig:c4}

\end{figure}
\clearpage
\section{Additional Experiments}\label{sec:sup_tests}
\setcounter{table}{0}
\setcounter{figure}{0}

\subsection{EVNet Variants}\label{sec:sup_variant}

\begin{table}[h]
\centering
\caption{\textbf{Detailed results for EVResNet50 variants, including ablations, on brain-alignment metrics.}  BrainScore~\cite{SchrimpfKubilius2018BrainScore, Schrimpf2020integrative} V1 alignment scores and shape bias~\cite{geirhos2022imagenettrained} for all EVResNet50 variants. ResNet50 and VOneResNet50 included for reference but not in the comparison. Values indicate mean \(\pm\) SD (\(n=2\) seeds).}
\label{tab:variant_brainscore}
\scalebox{0.8}{
\begin{tabular}{lcccccccccc}
\toprule
 & & & \multicolumn{7}{c}{V1 Response Properties} & \\ \cmidrule{4-10} 
 & V1 & V1 Resp. & Orient. & SF & RF & Surround & Texture & Resp. & Resp. & Shape\\
Model & Predict. & Prop. & Tuning & Tuning & Size & Mod. & Mod. & Select. & Magn. & Bias [\%]\\ \midrule
\textcolor{gray}{ResNet50} & \textcolor{gray}{.271} & \textcolor{gray}{.637} & \textcolor{gray}{.822} & \textcolor{gray}{.754} & \textcolor{gray}{.214} & \textcolor{gray}{.389} & \textcolor{gray}{.792} & \textcolor{gray}{.621} & \textcolor{gray}{.865} & \textcolor{gray}{18.8} \\
 & \textcolor{gray}{\(^{\pm.002}\)} & \textcolor{gray}{\(^{\pm.008}\)} & \textcolor{gray}{\(^{\pm.027}\)} & \textcolor{gray}{\(^{\pm.026}\)} & \textcolor{gray}{\(^{\pm.002}\)} & \textcolor{gray}{\(^{\pm.023}\)} & \textcolor{gray}{\(^{\pm.028}\)} & \textcolor{gray}{\(^{\pm.010}\)} & \textcolor{gray}{\(^{\pm.012}\)} & \textcolor{gray}{\(^{\pm1.2}\)} \\
\textcolor{gray}{VOneResNet50} & \textcolor{gray}{.375} & \textcolor{gray}{.754} & \textcolor{gray}{.859} & \textcolor{gray}{.969} & \textcolor{gray}{.482} & \textcolor{gray}{.373} & \textcolor{gray}{.919} & \textcolor{gray}{.792} & \textcolor{gray}{.884} & \textcolor{gray}{31.6} \\
 & \textcolor{gray}{\(^{\pm.002}\)} & \textcolor{gray}{\(^{\pm.006}\)} & \textcolor{gray}{\(^{\pm.005}\)} & \textcolor{gray}{\(^{\pm.001}\)} & \textcolor{gray}{\(^{\pm.041}\)} & \textcolor{gray}{\(^{\pm.003}\)} & \textcolor{gray}{\(^{\pm.004}\)} & \textcolor{gray}{\(^{\pm.003}\)} & \textcolor{gray}{\(^{\pm.002}\)} & \textcolor{gray}{\(^{\pm1.2}\)} \\ 
EVResNet50 & .364 & .826 & .854 & .950 & .726 & .614 & .916 & .781 & .933 & 48.9 \\
 & \(^{\pm\text{.000}}\) & \(^{\pm\text{.000}}\) & \(^{\pm\text{.009}}\) & \(^{\pm\text{.000}}\) & \(^{\pm\text{.001}}\) & \(^{\pm\text{.004}}\) & \(^{\pm\text{.001}}\) & \(^{\pm\text{.000}}\) & \(^{\pm\text{.000}}\) & \(^{\pm\text{2.4}}\) \\
\(-\) P Cells & .350 & \textbf{.845} & .854 & .945 & \textbf{.738} & \textbf{.737} & .910 & .764 & \textbf{.965} & \textbf{77.8} \\
& \(^{\pm\text{.006}}\) & \(^{\pm\textbf{.001}}\) & \(^{\pm\text{.002}}\) & \(^{\pm\text{.000}}\) & \(^{\pm\textbf{.006}}\) & \(^{\pm\textbf{.001}}\) & \(^{\pm\text{.007}}\) & \(^{\pm\text{.001}}\) & \(^{\pm\textbf{.005}}\) & \(^{\pm\textbf{1.7}}\) \\
\(-\) M Cells & .368 & .763 & .858 & .950 & .527 & .393 & .906 & .781 & .926 & 49.9 \\
 & \(^{\pm\text{.006}}\) & \(^{\pm\text{.000}}\) & \(^{\pm\text{.004}}\) & \(^{\pm\text{.000}}\) & \(^{\pm\text{.000}}\) & \(^{\pm\text{.007}}\) & \(^{\pm\text{.006}}\) & \(^{\pm\text{.004}}\) & \(^{\pm\text{.000}}\) & \(^{\pm\text{1.0}}\) \\
\(-\) Light Adapt. & .364 & .826 & .859 & \textbf{.951} & .720 & .630 & .908 & .780 & .936 & 70.0 \\
 & \(^{\pm\text{.001}}\) & \(^{\pm\text{.001}}\) & \(^{\pm\text{.002}}\) & \(^{\pm\textbf{.000}}\) & \(^{\pm\text{.006}}\) & \(^{\pm\text{.008}}\) & \(^{\pm\text{.009}}\) & \(^{\pm\text{.005}}\) & \(^{\pm\text{.006}}\) & \(^{\pm\text{3.3}}\) \\
\(-\) Contrast Norm. & \textbf{.374} & .768 & \textbf{.868} & .936 & .562 & .370 & \textbf{.921} & \textbf{.784} & .938 & 48.0 \\
 & \(^{\pm\textbf{.007}}\) & \(^{\pm\text{.002}}\) & \(^{\pm\textbf{.002}}\) & \(^{\pm\text{.001}}\) & \(^{\pm\text{.008}}\) & \(^{\pm\text{.001}}\) & \(^{\pm\textbf{.008}}\) & \(^{\pm\textbf{.005}}\) & \(^{\pm\text{.006}}\) & \(^{\pm\text{2.8}}\) \\ \bottomrule
\end{tabular}}
\end{table}

\begin{table}[h!]
\caption{\textbf{Detailed results for EVResNet50 variants, including ablations, on common corruption categories.} Clean and corrupted top-1 accuracies averaged across corruptions types for all EVResNet50 variants. ResNet50 and VOneResNet50 included for reference but not in the comparison. Values indicate mean \(\pm\) SD (\(n=2\) seeds).}
\label{tab:variant_corruptions}
\centering
\begin{tabular}{lcccccc}
\toprule
 & & \multicolumn{4}{c}{Corruption Types} & \\ \cmidrule(lr){3-6}
 & Mean & Noise & Blur & Weather & Digital & Clean \\
Model & {[}\%{]} & {[}\%{]} & {[}\%{]} & {[}\%{]} & {[}\%{]} & {[}\%{]} \\ \midrule
\textcolor{gray}{ResNet50} & \textcolor{gray}{38.8\scriptsize{\(\pm\)0.5}} & \textcolor{gray}{29.2\scriptsize{\(\pm\)0.6}} & \textcolor{gray}{34.6\scriptsize{\(\pm\)0.4}} & \textcolor{gray}{36.1\scriptsize{\(\pm\)0.5}} & \textcolor{gray}{49.5\scriptsize{\(\pm\)0.6}} & \textcolor{gray}{75.4\scriptsize{\(\pm\)0.1}} \\
\textcolor{gray}{VOneResNet50} & \textcolor{gray}{40.4\scriptsize{\(\pm\)0.1}} & \textcolor{gray}{35.9\scriptsize{\(\pm\)0.5}} & \textcolor{gray}{34.8\scriptsize{\(\pm\)0.1}} & \textcolor{gray}{32.6\scriptsize{\(\pm\)0.1}} & \textcolor{gray}{52.2\scriptsize{\(\pm\)0.1}} & \textcolor{gray}{72.9\scriptsize{\(\pm\)0.1}} \\
EVResNet50 & 41.9\scriptsize{\(\pm\)0.2} & 39.6\scriptsize{\(\pm\)0.3} & 37.5\scriptsize{\(\pm\)0.1} & 30.6\scriptsize{\(\pm\)0.2} & 53.5\scriptsize{\(\pm\)0.1} & 70.4\scriptsize{\(\pm\)0.1}\\
\(-\) P Cells & 34.9\scriptsize{\(\pm\)0.1} & \textbf{42.7}\scriptsize{\(\boldsymbol{\pm}\)\textbf{0.1}} & 26.8\scriptsize{\(\pm\)0.1} & 19.6\scriptsize{\(\pm\)0.1} & 45.7\scriptsize{\(\pm\)0.1} & 60.7\scriptsize{\(\pm\)0.1} \\
\(-\) M Cells & 42.1\scriptsize{\(\pm\)0.1} & 41.1\scriptsize{\(\pm\)0.2} & 37.7\scriptsize{\(\pm\)0.0} & 30.6\scriptsize{\(\pm\)0.1} & 53.3\scriptsize{\(\pm\)0.1} & 70.3\scriptsize{\(\pm\)0.1} \\
\(-\) Light Adaptation & 40.4\scriptsize{\(\pm\)0.2} & 35.2\scriptsize{\(\pm\)0.0} & \textbf{37.9}\scriptsize{\(\boldsymbol{\pm}\)\textbf{0.2}} & 29.2\scriptsize{\(\pm\)0.2} & 52.2\scriptsize{\(\pm\)0.1} & 69.8\scriptsize{\(\pm\)0.2} \\
\(-\) Contrast Norm. & 41.7\scriptsize{\(\pm\)0.4} & 39.2\scriptsize{\(\pm\)1.1} & 37.6\scriptsize{\(\pm\)0.4} & 30.4\scriptsize{\(\pm\)0.4} & 53.4\scriptsize{\(\pm\)0.1} & 70.7\scriptsize{\(\pm\)0.0} \\
\(-\) Subcort. Noise  & 41.9\scriptsize{\(\pm\)0.2} & 39.9\scriptsize{\(\pm\)0.5} & 35.8\scriptsize{\(\pm\)0.1} & 32.2\scriptsize{\(\pm\)0.5} & 53.8\scriptsize{\(\pm\)0.0} & \textbf{72.6}\scriptsize{\(\boldsymbol{\pm}\)\textbf{0.2}} \\
\(-\) VOneBlock & \textbf{42.7}\scriptsize{\(\boldsymbol{\pm}\)\textbf{0.2}} & 41.2\scriptsize{\(\pm\)0.9} & 37.4\scriptsize{\(\pm\)0.2} & \textbf{32.6}\scriptsize{\(\boldsymbol{\pm}\)\textbf{0.1}} & \textbf{54.2}\scriptsize{\(\boldsymbol{\pm}\)\textbf{0.1}} & 71.6\scriptsize{\(\pm\)0.1} \\
\(+\) LGN-V2 Conn. & 42.1\scriptsize{\(\pm\)0.1} & 40.0\scriptsize{\(\pm\)0.1} & 37.6\scriptsize{\(\pm\)0.3} & 30.8\scriptsize{\(\pm\)0.1} & 53.8\scriptsize{\(\pm\)0.0} & 70.7\scriptsize{\(\pm\)0.1} \\ \bottomrule
\end{tabular}
\end{table}

\begin{table}[h]
\centering
\caption{\textbf{Detailed results for EVResNet50 variants, including ablations, on domain-shift accuracy.} Top-1 accuracies on ImageNet-\{Cartoon, Drawing, R, Sketch, Stylized\(_{16}\)\} for all EVResNet50 variants. ResNet50 and VOneResNet50 included for reference but not in the comparison. Values indicate mean \(\pm\) SD (\(n=2\) seeds).}
\label{tab:variant_ood}
\begin{tabular}{lcccccc}
\toprule
 & Mean & Cartoon & Drawing & R & Sketch & Stylized\(_{16}\) \\
Model & {[}\%{]} & {[}\%{]} & {[}\%{]} & {[}\%{]} & {[}\%{]} & {[}\%{]} \\ \midrule
\textcolor{gray}{ResNet50} & \textcolor{gray}{33.4\scriptsize{\(\pm\)0.2}} & \textcolor{gray}{51.2\scriptsize{\(\pm\)0.7}} & \textcolor{gray}{20.9\scriptsize{\(\pm\)0.6}} & \textcolor{gray}{35.4\scriptsize{\(\pm\)0.1}} & \textcolor{gray}{23.3\scriptsize{\(\pm\)0.1}} & \textcolor{gray}{36.3\scriptsize{\(\pm\)1.2}} \\
\textcolor{gray}{VOneResNet50} & \textcolor{gray}{37.1\scriptsize{\(\pm\)0.4}} & \textcolor{gray}{55.5\scriptsize{\(\pm\)0.2}} & \textcolor{gray}{30.5\scriptsize{\(\pm\)0.4}} & \textcolor{gray}{37.5\scriptsize{\(\pm\)0.1}} & \textcolor{gray}{23.1\scriptsize{\(\pm\)0.3}} & \textcolor{gray}{38.8\scriptsize{\(\pm\)1.1}} \\
EVResNet50 & 38.1\scriptsize{\(\pm\)0.3} & 57.1\scriptsize{\(\pm\)0.3} & 33.9\scriptsize{\(\pm\)0.2} & 38.1\scriptsize{\(\pm\)0.2} & 22.7\scriptsize{\(\pm\)0.2} & 38.6\scriptsize{\(\pm\)1.1} \\
\(-\) P Cells & 33.6\scriptsize{\(\pm\)0.0} & 46.3\scriptsize{\(\pm\)0.2} & 20.1\scriptsize{\(\pm\)0.2} & 36.3\scriptsize{\(\pm\)0.3} & \textbf{24.7}\scriptsize{\(\boldsymbol{\pm}\)\textbf{0.1}} & \textbf{40.7}\scriptsize{\(\boldsymbol{\pm}\)\textbf{0.3}} \\
\(-\) M Cells & \textbf{38.2}\scriptsize{\(\boldsymbol{\pm}\)\textbf{0.2}} & 57.0\scriptsize{\(\pm\)0.2} & 34.4\scriptsize{\(\pm\)0.5} & 38.0\scriptsize{\(\pm\)0.0} & 22.8\scriptsize{\(\pm\)0.2} & 38.9\scriptsize{\(\pm\)0.9} \\
\(-\) Light Adaptation & 37.4\scriptsize{\(\pm\)0.2} & 56.2\scriptsize{\(\pm\)0.2} & 34.1\scriptsize{\(\pm\)0.5} & 37.4\scriptsize{\(\pm\)0.1} & 21.3\scriptsize{\(\pm\)0.4} & 38.1\scriptsize{\(\pm\)1.1} \\
\(-\) Contrast Norm. & 38.0\scriptsize{\(\pm\)0.1} & 56.9\scriptsize{\(\pm\)0.2} & 33.7\scriptsize{\(\pm\)0.4} & 38.1\scriptsize{\(\pm\)0.3} & 22.7\scriptsize{\(\pm\)0.6} & 38.6\scriptsize{\(\pm\)0.9} \\
\(-\) Subcort. Noise & 37.4\scriptsize{\(\pm\)0.1} & 56.5\scriptsize{\(\pm\)0.0} & 31.0\scriptsize{\(\pm\)0.3} & 37.6\scriptsize{\(\pm\)0.0} & 23.0\scriptsize{\(\pm\)0.1} & 38.7\scriptsize{\(\pm\)0.1} \\
\(-\) VOneBlock & \textbf{38.2}\scriptsize{\(\boldsymbol{\pm}\)\textbf{0.1}} & \textbf{57.2}\scriptsize{\(\boldsymbol{\pm}\)\textbf{0.0}} & \textbf{34.7}\scriptsize{\(\boldsymbol{\pm}\)\textbf{0.3}} & \textbf{38.2}\scriptsize{\(\boldsymbol{\pm}\)\textbf{0.2}} & 23.0\scriptsize{\(\pm\)0.3} & 38.1\scriptsize{\(\pm\)0.6} \\
\(+\) LGN-V2 Conn. & 38.3\scriptsize{\(\pm\)0.2} & 57.0\scriptsize{\(\pm\)0.2} & 34.5\scriptsize{\(\pm\)0.2} & 38.0\scriptsize{\(\pm\)0.2} & 22.8\scriptsize{\(\pm\)0.2} & 39.0\scriptsize{\(\pm\)0.5} \\ \bottomrule
\end{tabular}
\end{table}

\begin{table}[h!]
\centering
\caption{\textbf{Detailed results for EVResNet50 variants, including ablations, on adversarial robustness.} Top-1 accuracies for all EVResNet50 variants on limited adversarial set.  ResNet50 and VOneResNet50 included for reference but not in the comparison. Values indicate mean \(\pm\) SD (\(n=2\) seeds).}
\label{tab:variant_adversarial}
\begin{tabular}{lccccccc}
\toprule
 & & \multicolumn{2}{c}{\(\|\delta\|_\infty\)} & \multicolumn{2}{c}{\(\|\delta\|_2\)} & \multicolumn{2}{c}{\(\|\delta\|_1\)} \\ \cmidrule(lr){3-4} \cmidrule(lr){5-6} \cmidrule(lr){7-8}
 & Mean & \(\frac{1}{1020}\) & \(\frac{1}{255}\) & 0.15 & 0.6 & 40 & 160 \\
Model & {[}\%{]} & {[}\%{]} & {[}\%{]} & {[}\%{]} & {[}\%{]} & {[}\%{]} & {[}\%{]} \\ \midrule
\textcolor{gray}{ResNet50} & \textcolor{gray}{16.4} & \textcolor{gray}{23.4} & \textcolor{gray}{0.4} & \textcolor{gray}{37.2} & \textcolor{gray}{1.8} & \textcolor{gray}{33.6} & \textcolor{gray}{1.7} \\
& \textcolor{gray}{\(^{\pm0.5}\)} & \textcolor{gray}{\(^{\pm0.8}\)} & \textcolor{gray}{\(^{\pm0.0}\)} & \textcolor{gray}{\(^{\pm1.0}\)} & \textcolor{gray}{\(^{\pm0.2}\)} & \textcolor{gray}{\(^{\pm0.7}\)} & \textcolor{gray}{\(^{\pm0.2}\)} \\
\textcolor{gray}{VOneResNet50} & \textcolor{gray}{50.5} & \textcolor{gray}{62.6} & \textcolor{gray}{30.4} & \textcolor{gray}{66.2} & \textcolor{gray}{42.3} & \textcolor{gray}{64.5} & \textcolor{gray}{37.3} \\
& \textcolor{gray}{\(^{\pm0.1}\)} & \textcolor{gray}{\(^{\pm0.4}\)} & \textcolor{gray}{\(^{\pm0.3}\)} & \textcolor{gray}{\(^{\pm0.3}\)} & \textcolor{gray}{\(^{\pm0.2}\)} & \textcolor{gray}{\(^{\pm0.3}\)} & \textcolor{gray}{\(^{\pm0.6}\)} \\
EVResNet50 & 53.8 & 62.7 & 38.8 & 65.1 & 48.0 & 64.0 & 44.5 \\
& \(^{\pm\text{0.2}}\) & \(^{\pm\text{0.2}}\) & \(^{\pm\text{0.6}}\) & \(^{\pm\text{0.0}}\) & \(^{\pm\text{0.3}}\) & \(^{\pm\text{0.2}}\) & \(^{\pm\text{0.4}}\) \\
\(-\) P Cells & 46.7 & 53.4 & 33.4 & 55.6 & 42.4 & 55.2 & 40.3 \\
& \(^{\pm\text{0.2}}\) & \(^{\pm\text{0.3}}\) & \(^{\pm\text{0.1}}\) & \(^{\pm\text{0.5}}\) & \(^{\pm\text{0.2}}\) & \(^{\pm\text{0.4}}\) & \(^{\pm\text{0.1}}\) \\
\(-\) M Cells & \textbf{54.0} & 62.6 & 40.1 & 64.8 & \textbf{48.4 }& 64.1 & 44.0 \\
& \(^{\pm\textbf{0.1}}\) & \(^{\pm\text{0.3}}\) & \(^{\pm\text{0.2}}\) & \(^{\pm\textbf{0.1}}\) & \(^{\pm\text{0.2}}\) & \(^{\pm\text{0.1}}\) & \(^{\pm\text{0.7}}\) \\
\(-\) Light Adapt. & 55.1 & \textbf{62.8} & \textbf{42.1} & 65.0 & 50.3 & 64.0 & \textbf{46.3} \\
& \(^{\pm\text{0.5}}\) & \(^{\pm\textbf{0.5}}\) & \(^{\pm\textbf{0.4}}\) & \(^{\pm\text{0.7}}\) & \(^{\pm\text{0.9}}\) & \(^{\pm\text{0.2}}\) & \(^{\pm\textbf{0.3}}\) \\
\(-\) Contrast Norm. & 53.3 & 62.7 & 38.2 & \textbf{65.2} & 47.3 & 64.3 & 43.5 \\
& \(^{\pm\text{0.1}}\) & \(^{\pm\text{0.2}}\) & \(^{\pm\text{0.4}}\) & \(^{\pm\textbf{0.3}}\) & \(^{\pm\text{0.2}}\) & \(^{\pm\text{0.3}}\) & \(^{\pm\text{0.0}}\) \\
\(-\) Subcort. Noise & 48.5 & 60.5 & 28.2 & 64.2 & 39.6 & 62.7 & 35.8 \\
& \(^{\pm\text{0.1}}\) & \(^{\pm\text{0.1}}\) & \(^{\pm\text{0.8}}\) & \(^{\pm\text{0.1}}\) & \(^{\pm\text{0.3}}\) & \(^{\pm\text{0.1}}\) & \(^{\pm\text{0.3}}\) \\
\(-\) VOneBlock & 51.8 & 62.2 & 34.3 & \textbf{65.2} & 44.8 & 63.9 & 40.4 \\
& \(^{\pm\text{0.5}}\) & \(^{\pm\text{0.3}}\) & \(^{\pm\text{0.6}}\) & \(^{\pm\textbf{0.8}}\) & \(^{\pm\text{0.2}}\) & \(^{\pm\text{0.9}}\) & \(^{\pm\text{0.4}}\) \\
\(+\) LGN-V2 Conn. & 53.9 & \textbf{62.8} & 38.7 & 65.0 & \textbf{48.4} & \textbf{64.5} & 44.2 \\
& \(^{\pm\text{0.2}}\) & \(^{\pm\textbf{0.3}}\) & \(^{\pm\text{0.1}}\) & \(^{\pm\text{0.8}}\) & \(^{\pm\textbf{0.2}}\) & \(^{\pm\textbf{0.2}}\) & \(^{\pm\text{0.2}}\) \\ \bottomrule
\end{tabular}
\end{table}

\clearpage
\subsection{EVNet Backend Generalization}\label{sec:sup_backend}

Similarly to the results obtained with EVResNet50, the EVEfficientNet-B0 and EVCORnet-Z models consistently outperform their corresponding base model across most corruption categories, as well as in mean corruption accuracy, as shown in Table~\ref{tab:backend_corruptions}. However, this improvement comes with a greater relative drop in clean image accuracy compared to the ResNet50-based models. This steeper drop likely reflects architectural differences in sensitivity to input statistics. EfficientNet-B0 employs compound scaling and aggressive architecture search to optimize performance specifically for standard ImageNet inputs~\cite{tan2020efficientnet}, making it more susceptible to deviations introduced by our biologically inspired preprocessing. In contrast, ResNet50, with its more generic design appears to be more adaptable to altered input distribution. Similarly, compared to ResNet50, the compact architecture of CORnet-Z exhibits a lower degree of feature redundancy which, when coupled with a mismatch between the inductive biases imposed by the front-end and those the network was designed to exploit, can limit its flexibility to adapt to the upstream processing. % Confirmar
When evaluated on domain shift datasets, both EVEfficientNet-B0 and EVCORnet-Z surpass their base models on most benchmarks, as reported in Table~\ref{tab:backend_ood}, with the only exception being ImageNet-Sketch, mimicking the same pattern as observed with the EVResNet50. Both EVEfficientNet-B0 and EVCORnet-Z exhibit substantial improvements across all norm constraints (Table~\ref{tab:backend_adversarial}) and, when aggregated into the Robustness Score (Table~\ref{tab:backend_robust}), EVNets consistently surpass their respective base architectures, reinforcing the effectiveness of back-end generalization.

\begin{table}[h!]
\centering
\caption{\textbf{EVNets outperforms base models on most image corruption types and on mean corruption accuracy, across different backend architectures.} Clean and corrupted top-1 accuracies averaged across severities and corruptions for EfficientNet-B0, EVEfficientNet-B0, CORnet-Z and EVCORnet-Z. Values indicate mean \(\pm\) SD (\(n=2\) seeds).}
\label{tab:backend_corruptions}
\begin{tabular}{lcccccc}
\toprule
 & & \multicolumn{4}{c}{Corruption Types} & \\ \cmidrule(lr){3-6}
 & Mean & Noise & Blur & Weather & Digital & Clean \\
 Model & {[}\%{]} & {[}\%{]} & {[}\%{]} & {[}\%{]} & {[}\%{]} & {[}\%{]} \\ \midrule
EfficientNet-B0 & 30.3\scriptsize{\(\pm\)0.4} & 18.7\scriptsize{\(\pm\)0.2} & 26.6\scriptsize{\(\pm\)0.0} & \textbf{29.1\scriptsize{\(\pm\)0.3}} & 40.8\scriptsize{\(\pm\)1.2} & \textbf{68.1\scriptsize{\(\pm\)0.1}} \\
EVEfficientNet-B0 & \textbf{34.1\scriptsize{\(\pm\)0.0}} & \textbf{30.7\scriptsize{\(\pm\)0.2}} & \textbf{30.5\scriptsize{\(\pm\)0.3}} & 24.3\scriptsize{\(\pm\)0.1} & \textbf{45.0\scriptsize{\(\pm\)0.1}} & 61.4\scriptsize{\(\pm\)0.4} \\ \midrule
CORnet-Z & 18.0\scriptsize{\(\pm\)0.0} & 6.4\scriptsize{\(\pm\)0.1} & 17.0\scriptsize{\(\pm\)0.0} & \textbf{12.4\scriptsize{\(\pm\)0.3}} & 29.1\scriptsize{\(\pm\)0.1} & \textbf{53.2\scriptsize{\(\pm\)0.1}} \\ 
EVCORnet-Z & \textbf{21.3\scriptsize{\(\pm\)0.0}} & \textbf{20.0\scriptsize{\(\pm\)0.0}} & \textbf{18.2\scriptsize{\(\pm\)0.1}} & 11.6\scriptsize{\(\pm\)0.0} & \textbf{30.4\scriptsize{\(\pm\)0.0}} & 44.7\scriptsize{\(\pm\)0.0} \\ \bottomrule
\end{tabular}
\end{table}

\begin{table}[h!]
\centering
\caption{\textbf{EVNets outperforms base models on most OOD datasets and on mean domain shift accuracy, across different backend architectures.} Top-1 accuracies on ImageNet-\{Cartoon, Drawing, R, Sketch, Stylized\(_{16}\)\} for EfficientNet-B0, EVEfficientNet-B0, CORnet-Z and EVCORnet-Z. Values indicate mean \(\pm\) SD (\(n=2\) seeds).}
\label{tab:backend_ood}
\begin{tabular}{lcccccc}
\toprule
 & Mean & Cartoon & Drawing & R & Sketch & Stylized\(_{16}\) \\
Model & {[}\%{]} & {[}\%{]} & {[}\%{]} & {[}\%{]} & {[}\%{]} & {[}\%{]} \\ \midrule
EfficientNet-B0 & 28.9\scriptsize{\(\pm\)0.2} & 40.2\scriptsize{\(\pm\)0.4} & 17.1\scriptsize{\(\pm\)0.5} & 29.9\scriptsize{\(\pm\)0.2} & \textbf{17.3\scriptsize{\(\pm\)0.5}} & 40.0\scriptsize{\(\pm\)0.7} \\
EVEfficientNet-B0 & \textbf{33.5\scriptsize{\(\pm\)0.2}} & \textbf{49.1\scriptsize{\(\pm\)0.0}} & \textbf{28.2\scriptsize{\(\pm\)0.2}} & \textbf{31.5\scriptsize{\(\pm\)0.2}} & 16.8\scriptsize{\(\pm\)0.3} & \textbf{41.8\scriptsize{\(\pm\)1.0}} \\ \midrule
CORnet-Z & 19.5\scriptsize{\(\pm\)0.2} & 30.9\scriptsize{\(\pm\)0.2} & 12.5\scriptsize{\(\pm\)0.3} & 21.0\scriptsize{\(\pm\)0.2} & \textbf{9.5\scriptsize{\(\pm\)0.1}} & 23.8\scriptsize{\(\pm\)1.8} \\
EVCORnet-Z & \textbf{21.1\scriptsize{\(\pm\)0.3}} & \textbf{32.9\scriptsize{\(\pm\)0.1}} & \textbf{16.2\scriptsize{\(\pm\)0.2}} & \textbf{21.3\scriptsize{\(\pm\)0.0}} & 9.0\scriptsize{\(\pm\)0.1} & \textbf{26.0\scriptsize{\(\pm\)1.1}} \\ \bottomrule
\end{tabular}
\end{table}

\begin{table}[h!]
\centering
\caption{\textbf{EVNets outperforms base models on most adversarial perturbations and on mean adversarial robustness, across different backend architectures.} Top-1 accuracies for EfficientNet-B0, EVEfficientNet-B0, CORnet-Z and EVCORnet-Z. Values indicate mean \(\pm\) SD (\(n=2\) seeds).}
\label{tab:backend_adversarial}
\begin{tabular}{lccccccc}
\toprule
 & & \multicolumn{2}{c}{\(\|\delta\|_\infty\)} & \multicolumn{2}{c}{\(\|\delta\|_2\)} & \multicolumn{2}{c}{\(\|\delta\|_1\)} \\ \cmidrule(lr){3-4} \cmidrule(lr){5-6} \cmidrule(lr){7-8}
 & Mean & \(\frac{1}{1020}\) & \(\frac{1}{255}\) & 0.15 & 0.6 & 40 & 160 \\
Model & {[}\%{]} & {[}\%{]} & {[}\%{]} & {[}\%{]} & {[}\%{]} & {[}\%{]} & {[}\%{]} \\ \midrule
EfficientNet-B0 & 20.9 & 35.0 & 2.0 & 43.5 & 5.1 & 36.2 & 3.3 \\
& \(^{\pm\text{0.1}}\) & \(^{\pm\text{0.6}}\) & \(^{\pm\text{0.1}}\) & \(^{\pm\text{0.3}}\) & \(^{\pm\text{0.1}}\) & \(^{\pm\text{0.2}}\) & \(^{\pm\text{0.1}}\) \\
EVEfficientNet-B0 & \textbf{45.6} & \textbf{53.2} & \textbf{31.1} & \textbf{55.8} & \textbf{40.8} & \textbf{55.0} & \textbf{37.7} \\
& \(^{\pm\textbf{0.5}}\) & \(^{\pm\textbf{0.5}}\) & \(^{\pm\textbf{0.2}}\) & \(^{\pm\textbf{0.6}}\) & \(^{\pm\textbf{0.5}}\) & \(^{\pm\textbf{0.6}}\) & \(^{\pm\textbf{0.6}}\) \\ \midrule
CORnet-Z & 17.6 & 24.4 & 0.7 & 34.8 & 5.4 & 34.7 & 5.5 \\
& \(^{\pm\text{0.3}}\) & \(^{\pm\text{0.7}}\) & \(^{\pm\text{0.0}}\) & \(^{\pm\text{0.6}}\) & \(^{\pm\text{0.1}}\) & \(^{\pm\text{0.6}}\) & \(^{\pm\text{0.0}}\) \\
EVCORnet-Z & \textbf{31.1} & \textbf{36.7} & \textbf{19.0} & \textbf{38.8} & \textbf{26.9} & \textbf{39.0} & \textbf{26.2} \\
& \(^{\pm\text{0.6}}\) & \(^{\pm\text{0.9}}\) & \(^{\pm\text{0.8}}\) & \(^{\pm\text{0.6}}\) & \(^{\pm\text{0.4}}\) & \(^{\pm\text{0.6}}\) & \(^{\pm\text{0.4}}\) \\ \bottomrule
\end{tabular}
\end{table}

\begin{table}[h!]
\centering
\caption{\textbf{EVNets outperforms base models on Robustness Score, across different backend architectures.} Robustness Score, clean and perturbed top-1 accuracies for EfficientNet-B0, EVEfficientNet-B0, CORnet-Z and EVCORnet-Z. Values indicate mean \(\pm\) SD (\(n=2\) seeds).}
\label{tab:backend_robust}
\begin{tabular}{lccccc}
\toprule
 &  & \multicolumn{3}{c}{Perturbations} & \\ \cmidrule{3-5}
 & Robust. Score* & Adversarial* & Corrupt. & Domain Shift & Clean \\
Model & {[}\%{]} & {[}\%{]} & {[}\%{]} & {[}\%{]} & {[}\%{]} \\ \midrule
EfficientNet-B0 & 26.7\scriptsize{\(\pm\)0.1} & 20.9\scriptsize{\(\pm\)0.1} & 30.3\scriptsize{\(\pm\)0.4} & 28.9\scriptsize{\(\pm\)0.2} & \textbf{68.1\scriptsize{\(\pm\)0.1}} \\
EVEfficientNet-B0 & \textbf{39.7\scriptsize{\(\pm\)0.3}} & \textbf{45.6\scriptsize{\(\pm\)0.4}} & \textbf{34.1\scriptsize{\(\pm\)0.0}} & \textbf{33.5\scriptsize{\(\pm\)0.1}} & 61.4\scriptsize{\(\pm\)0.4} \\ \midrule
CORnet-Z & 18.4\scriptsize{\(\pm\)0.0} & 17.6\scriptsize{\(\pm\)0.3} & 18.0\scriptsize{\(\pm\)0.0} & 19.5\scriptsize{\(\pm\)0.2} & \textbf{53.2\scriptsize{\(\pm\)0.1}} \\
EVCORnet-Z & \textbf{24.5\scriptsize{\(\pm\)0.3}} & \textbf{31.1\scriptsize{\(\pm\)0.6}} & \textbf{21.3\scriptsize{\(\pm\)0.0}} & \textbf{21.1\scriptsize{\(\pm\)0.3}} & 44.7\scriptsize{\(\pm\)0.0} \\ \bottomrule
\end{tabular}
\end{table}

\subsection{EVNet Inference Ensembling}\label{sec:ensemble_inference}

\begin{figure}[h!]
\centering
\begin{tikzpicture}
\pgftext{\includegraphics[width=1.0\linewidth]{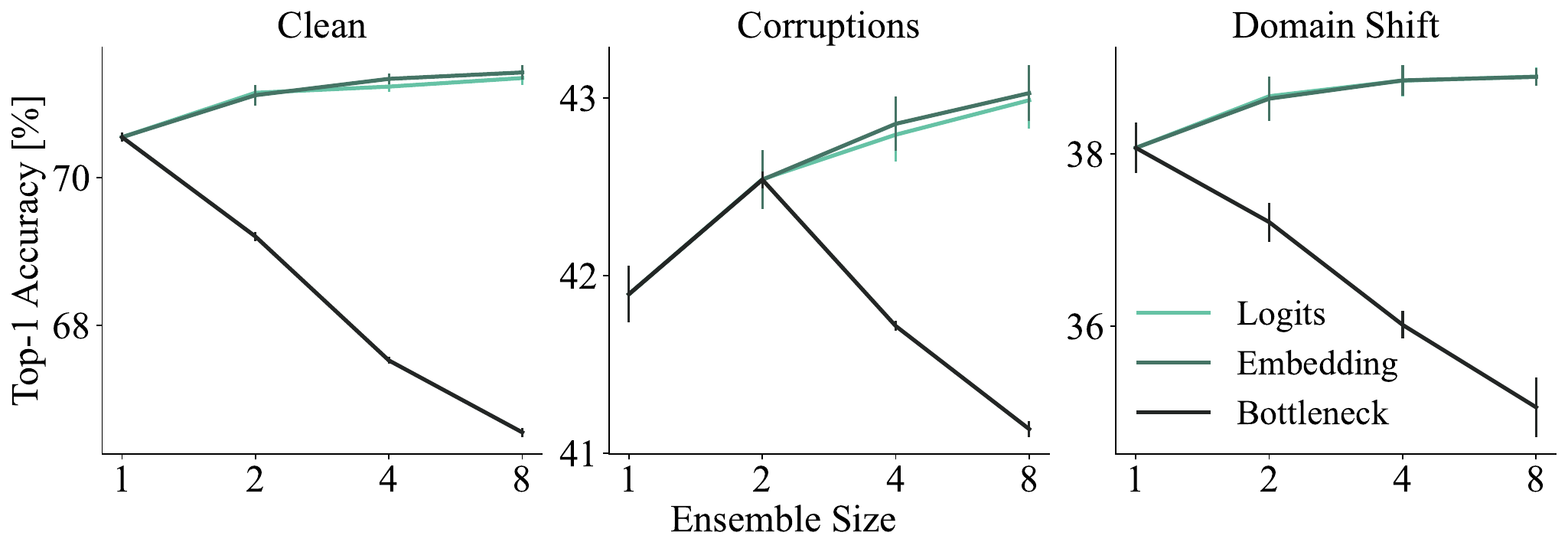}};
\end{tikzpicture}
\caption{\textbf{Averaging EVNet logits or final embeddings yields slight performance improvements, whereas averaging bottleneck activations degrades accuracy.} Accuracy under clean, corruption, and domain-shift images for EVResNet50 ensembles of varying sizes. ``Logits'' denotes ensembles averaged at the logit layer, ``Embeddings'' refers to averaging activations in \texttt{layer4} (prior to global average pooling), and ``Bottleneck'' indicates averaging immediately after the VOneBlock bottleneck. Lines indicate mean top-1 accuracy and error bars represent SD (\(n=3\) seeds)}\label{fig:ensemble_inference}
\end{figure}

To evaluate whether combining the stochastic activations of EVNets across multiple forward passes leads to cumulative performance gains, we conducted an ensemble analysis varying both ensemble size and the stage at which activations are aggregated. Specifically, we compared ensembles that averaged activations at three points in the network: (1) the logit layer, (2) the final embedding stage (\texttt{layer4}, before the global average pooling), and (3) immediately after the VOneBlock bottleneck. We found that averaging later representations at the embedding or logit level yielded marginal but consistent improvements across clean, corruption, and domain-shift evaluations (Fig.~\ref{fig:ensemble_inference}). In contrast, averaging activations after the bottleneck reduced performance, with the exception of a small performance gain by the two-model ensemble when evaluated on ImageNet-C. This degradation likely arises because the bottleneck lies immediately downstream of the noise-injection stage, and averaging at this point effectively diminishes the stochastic variability that the EVNet leverages during training. We did not evaluate adversarial performance in this setting, as generating adversarial samples already requires an ensemble of forward passes, and using an additional ensemble for evaluation would compound computational costs to an impractical level.

\clearpage
\section{Implementation Details}\label{sec:sup_implementation}
\setcounter{table}{0}
\setcounter{figure}{0}

\subsection{Grating Experiments}\label{sec:sup_gratings}

When tuning the SubcorticalBlock and when measuring its response properties, we presented 12 frames of drifting sine-wave gratings with phase shifts of 30 degrees in the interval [0, 360[ degrees. Grating orientation was set to horizontal and the diameter, SF and contrast was chosen to most accurately replicate the response-property studies used for tuning (see Table~\ref{tab:resp_prop} for the reference studies from which the stimulus set properties were taken). The background area not covered by the grating was set to 50\% gray. To characterize the response properties of the VOneBlock, we adopted the same experimental paradigm as detailed above. %In quantifying V1-like tuning, we followed established neurophysiological conventions \cite{DEVALOIS1982545}, computing the mean response (F0) for complex cells and the amplitude of the first harmonic (F1) for simple cells.

\subsection{SubcorticalBlock Implementation}\label{sec:sup_subcorticalblock_implementation}

To parameterize the fixed weights of the SubcorticalBlock, we developed a novel tuning strategy that optimizes alignment with average neuronal response properties of SF tuning, size tuning, and contrast sensitivity, using Bayesian optimization. Table~\ref{tab:subcortical_prop} shows the reference values and values obtained values for the six subcortical response properties. This procedure was applied independently to the P and M pathways within the SubcorticalBlock. While several hyperparameters were directly optimized, Gaussian kernel sizes were indirectly determined by computing the kernel size necessary to elicit 75\% of the their total integrated response. Specifically, this formulation was used with the surround Gaussian in the DoG kernel and with the Gaussian kernel of contrast normalization layer.

\paragraph{Light adaptation pooling size.} Because the primate visual system exhibits both global and local forms of light adaptation, we initially modeled luminance adaptation as a spatially local process, implemented via Gaussian filtering analogous to our contrast normalization layer. Interestingly, during Bayesian optimization, the learned filter radius consistently expanded to encompass nearly the entire image, suggesting that global rather than local adaptation better supported LGN response property prediciton. To reduce computational overhead, we therefore adopted a global luminance normalization strategy. % Reviewer 4 asked to clarify design choice of global light adaptation

\begin{table}[h]
\centering
\caption{\textbf{Reference and tuned response property values for the SubcorticalBlock.} Response property values specific to P and M cells used to tune the SubcorticalBlock, shown alongside reference values from the original studies from which they were sourced. Six response properties were used for tuning: center, surround, excitation and inhibition radii; suppression index; and saturation index.}
\label{tab:subcortical_prop}
\begin{tabular}{lcccc}
    \toprule
     & \multicolumn{2}{c}{Reference} & \multicolumn{2}{c}{SubcorticalBlock} \\ \cmidrule(lr){2-3} \cmidrule(lr){4-5}
    Resp. Property & P cells & M cells & P cells & M cells \\ \midrule
    Center Radius [deg]~\cite{Solomon338} & 0.042 & 0.063 & 0.041 & 0.064 \\
    Surround Radius [deg]~\cite{Solomon338} & 0.279 & 0.602 & 0.289 & 0.620 \\
    Excitation Radius [deg]~\cite{Solomon338} & 0.236 & 0.289 & 0.094 & 0.125 \\
    Inhibition Radius [deg]~\cite{Solomon338} & 0.564 & 0.869 & 0.226 & 0.609 \\
    Suppression Index~\cite{Solomon338} & 0.808 & 0.719 & 0.710 & 0.610 \\
    Saturation Index~\cite{RaghavanENEURO.0515-22.2023} & 0.095 &  0.365 & 0.200 & 0.410 \\ \bottomrule
\end{tabular}
\end{table}

\paragraph{Search space.} When defining the search space for each variable in our Bayesian optimization framework, our primary objective was to minimize the introduction of inductive biases by employing search spaces as broad as feasible. In many cases, this was straightforward --- for instance, we constrained parameters like the semisaturation constant, \(c_\text{50}\) to lie within physically meaningful bounds. However, for parameters such as the center and surround radii of the DoG filters, the radius of the Gaussian used in the contrast normalization layer, and the ratio of peak contrast sensitivity, we adopted a more heuristic approach. Specifically, we drew on values reported in the neuroscience literature to inform the bounds of the search space. For the DoG center and surround radii, we defined symmetric search intervals of centered around values reported in the reference study to which we aimed to maximize alignment~\cite{Solomon338}. Similarly, the bounds for the normalization radius were guided by reported relationships between the suppressive field and the surround Gaussian~\cite{Bonin2005-cu}. The same principle was applied to the contrast sensitivity ratio~\cite{CRONER19957}. Table~\ref{tab:params} provides a comprehensive overview of all parameters tuned, including the corresponding search bounds, the literature references used to guide their selection, where applicable, and the final hyperparameters.

\begin{table}[h]
\centering
\caption{\textbf{SubcorticalBlock hyperparameters and search space used for tuning.} Minimum (\(x_\text{min}\)) and maximum (\(x_\text{max}\)) bounds used in the Bayesian optimization for each hyperparameter of the SubcorticalBlock and hyperparameter optima obtained (\(x^\ast\)). Values describe center radius of the DoG (\(r_c\)); surround radius of the DoG (\(r_s\)); peak contrast sensitivity ratio (\(k_s / k_c\)); contrast normalization pooling radius (\(r_\text{CN}\)); semisaturation constant (\(c_{50}\)); contrast normalization exponent (\(n\)). While not obtained through optimization, the kernel sizes used (\(k_\text{DoG}\) and \(k_\text{CN}\)) are also presented. For a subset of these hyperparameters, literature references were used to inform the choice of search bounds.}
\label{tab:params}
\begin{tabular}{llccccccc}
    \toprule
     & & \multicolumn{3}{c}{P cells} & \multicolumn{3}{c}{M cells} & \\ \cmidrule(lr){3-5}\cmidrule(lr){6-8}
    Layer & \(x\) & \(x_\text{min}\) & \(x_\text{max}\) & \(x^\ast\) & \(x_\text{min}\) & \(x_\text{max}\) & \(x^\ast\) & Ref. \\ \midrule
    DoG & \(r_\text{c}\) [deg] & 0.034 & 0.050 & 0.047 & 0.050 & 0.76 & 0.76 & \cite{Solomon338} \\
    Conv. & \(r_\text{s}\) [deg] & 0.223 & 0.335 & 0.224 & 0.482 & 0.722 & 0.534 & \cite{Solomon338} \\
    & \(k_\text{s}/k_\text{c}\) & -0.068 & -0.003 & -0.12 & -0.037 & -0.002 & -0.004 & \cite{CRONER19957} \\
    & \(k_\text{DoG}\) & --- & --- & 19 & --- & --- & 33 & --- \\\midrule
    Contrast & \(r_\text{CN}\) [deg] & 0.140 & 0.419 & 0.419 & 0.301 & 0.903 & 0.902 & \cite{Bonin2005-cu} \\
    Norm. & \(c_\text{50}\) & 0.01 & 1.0 & 1.0 & 0.01 & 1.0 & 0.19 & --- \\
    & \(n\) & 0.01 & 1.0 & 1.0 & 0.01 & 1.0 & 0.81 & --- \\
    & \(k_\text{CN}\) & --- & --- & 43 & --- & --- & 69 & --- \\ \bottomrule
\end{tabular}
\end{table}

\paragraph{Bayesian optimization.}\label{sec:sup_bayesian_opt} 
We employed Bayesian optimization using the \texttt{gp\_minimize} function from the Scipy library~\cite{2020SciPy}. The optimization was performed over a defined parameter space for 640 evaluations, with 64 initial points generated using a Sobol sequence. The acquisition function was probabilistically selected among Lower Confidence Bound (LCB), Expected Improvement (EI), and Probability of Improvement (PI) at each iteration. The exploration-exploitation balance was controlled using \(\kappa=1.96\) for LCB and \(\xi=0.01\) for EI and PI.

\subsection{EVNet Variants}

For all EVNet variants, we re-estimated the scaling factor applied to the VOneBlock whenever it was included, and adjusted the V1 noise Fano factor such that the accumulated Fano factor was 1. Apart from these modifications, most variants were derived by simply performing the modifications described in previous sections, with the two exceptions detailed below.

\paragraph{Contrast normalization ablation.} Because the light adaptation and contrast normalization layers operate in close synchrony, removing the contrast normalization layer substantially destabilized training. In particular, the absence of contrast normalization caused activations within the SubcorticalBlock to explode, primarily due to excessively high responses from the light adaptation mechanism. This effect was most pronounced when image (or image crops) contained small, bright regions surrounded by dark backgrounds --- conditions that produce low mean activations but locally high responses in Equation ~\ref{eq:la}. To mitigate this, we modified the light adaptation layer’s mean computation to ignore pixel values below a threshold of \(\epsilon=0.05\), effectively preventing spurious amplification of isolated bright pixels.

\paragraph{LGN–V2 skip connection.} When incorporating the skip connections between the SubcorticalBlock and the VOneBlock bottleneck, we maintained a total channel dimensionality of 64 at the input to the backend model. Of these, 60 channels originated from the bottleneck output, while the remaining 4 channels were adapted activation maps from the SubcorticalBlock. Given that the VOneBlock operates with a stride of 4, we applied a \(5\times 5\) max-pooling operation with the same stride to the SubcorticalBlock activations prior to concatenation, ensuring spatial alignment and consistent feature scaling across pathways.

\subsection{Training Details}\label{sec:sup_training}

All models were trained on an internal cluster, using 48GB NVIDIA A40 GPUs with Python 3.11, PyTorch 2.2 with CUDA 11.7, taking roughly tree days to train.

\paragraph{Preprocessing.} During training, images were randomly horizontally flipped with a probability of 0.5, then resized and cropped to 224\(\times\)224 pixels. Images were normalized by subtracting and dividing by [0.5, 0.5, 0.5], with the exception of 
model that included the light adaptation layer of the SubcorticalBlock. During evaluation, images were resized to 256 pixels on the shorter side, followed by a center crop to 224\(\times\)224 pixels, and the same normalization was applied.

\paragraph{Loss function and optimization.} Models were trained using a cross-entropy loss between ground-truth labels and predicted logits, with label smoothing~\cite{7780677} of 0.1. When using the ResNet50 and CORnet-Z architectures, optimization was performed using stochastic gradient descent with momentum set to 0.9 and weight decay of \(5\times10^{-4}\). For EfficientNet-B0, we used RMSProp with a momentum of 0.9, smoothing constant of 0.9, and a denominator stability term of 1.0. Training was conducted for 50 epochs with a batch size of 256. We employed the 1-Cycle learning rate policy~\cite{smith2018}, where the learning rate was initialized at 4\% of the maximum learning rate, increased up to maximum at 30\% of the total training steps, and then annealed to \(4\times10^{-4}\%\) of the maximum following a cosine schedule. For the ResNet50, the maximum learning rate was set to 0.1; for the EfficientNet-B0, it was set to 0.256; and, for CORnet-Z, it was set to 0.05. When using PRIME~\cite{10.1007/978-3-031-19806-9_36}, we fine-tuned a standardly trained model for an additional 50 epochs using the same training protocol, except with a maximum learning rate of 0.01 reached at 10\% of the training schedule.

\end{document}